\documentclass[11pt]{article}

\usepackage[final]{acl}

\usepackage{times}
\usepackage{latexsym}

\usepackage[T1]{fontenc}

\usepackage[utf8]{inputenc}

\usepackage{microtype}

\usepackage{inconsolata}

\usepackage{graphicx}

\usepackage{hyperref}
\usepackage{url}
\usepackage{algorithm}

\usepackage{graphicx}
\usepackage{amsmath}
\usepackage{booktabs}
\usepackage{multirow}
\usepackage{enumitem}
\usepackage{algorithmicx}
\usepackage{geometry}  
\usepackage{xcolor}
\usepackage{color, colortbl}
\usepackage{caption}
\usepackage{adjustbox}
\usepackage{tcolorbox} 
\usepackage{todonotes}

\newcommand{\proposedmodel}{\texttt{SpatialMath}}
\newcommand{\proposedmodelfull}{\texttt{Spatial Comprehension-Infused Symbolic Reasoning Framework}}
\newcommand{\spc}{\texttt{SC}}
\newcommand{\spcfull}{\texttt{Spatial  Comprehension}}
\newcommand{\spcfulls}{\texttt{Spatial Comprehensions}}
\newcommand{\rcn}{\texttt{RC}}
\newcommand{\rcfull}{\texttt{Reasoning Chain}}
\newcommand{\newdataset}{\texttt{MATHVERSE-PLUS}}

\newcommand{\mcmfull}{\texttt{SpatialMath-SX:} Spatial Comprehension Core}
\newcommand{\mcm}{\texttt{SpatialMath-SX}}

\newcommand{\mrmfull}{\texttt{SpatialMath-RX:} Reasoning Infusion Core}
\newcommand{\mrm}{\texttt{SpatialMath-RX}}
\definecolor{LighterGreen}{rgb}{0.85,9,0.85}
\definecolor{LightRed}{rgb}{0.99,0.8,0.8}
\definecolor{backred}{RGB}{255, 190, 190}
\definecolor{backblue}{RGB}{210, 230, 250}

\newcommand\blfootnote[1]{%
  \begingroup
  \renewcommand\thefootnote{}\footnote{#1}%
  \addtocounter{footnote}{-1}%
  \endgroup
}
\title{SpatialMath: Spatial Comprehension-Infused Symbolic Reasoning for Mathematical Problem-Solving}

\author{\textbf{Ashutosh Bajpai\textsuperscript{1,*}},  \textbf{Akshat Bhandari\textsuperscript{1,*}},  \textbf{Akshay Nambi\textsuperscript{3}},  \textbf{Tanmoy Chakraborty\textsuperscript{1,2}}\\
\\
\textsuperscript{1}Indian Institute of Technology Delhi, India \\    
\textsuperscript{2}Indian Institute of Technology Abu Dhabi, UAE \\ 
\textsuperscript{3}Microsoft Research, India \\    
\textit{\{eez228482,tanchak\}@ee.iitd.ac.in}, \textit{akshatbhandari15@gmail.com}, \\
\textit{akshayn@microsoft.com}}    
  
\begin{document}
\maketitle
\begin{abstract}
Multimodal Small-to-Medium sized Language Models (MSLMs) have demonstrated strong capabilities in integrating visual and textual information but still face significant limitations in visual comprehension and mathematical reasoning, particularly in geometric problems with diverse levels of visual infusion. Current models struggle to accurately decompose intricate visual inputs and connect perception with structured reasoning, leading to suboptimal performance. To address these challenges, we propose \proposedmodel, a novel \proposedmodelfull\ designed to integrate spatial representations into structured symbolic reasoning chains. \proposedmodel\ employs a specialized perception module to extract spatially-grounded representations from visual diagrams, capturing critical geometric structures and spatial relationships. These representations are then methodically infused into symbolic reasoning chains, facilitating visual comprehension-aware structured reasoning. 
To this end, we introduce \newdataset, a novel dataset containing structured visual interpretations and step-by-step reasoning paths for vision-intensive mathematical problems. \proposedmodel\ significantly outperforms strong multimodal baselines, achieving up to 10 percentage points improvement over supervised fine-tuning with data augmentation in vision-intensive settings. Robustness analysis reveals that enhanced spatial representations directly improve reasoning accuracy, reinforcing the need for structured perception-to-reasoning pipelines in MSLMs.\blfootnote{\textsuperscript{*}Equal Contribution.}
\end{abstract}

\section{Introduction}
\label{sec:introduction}

Recent progress in generative AI has produced Multimodal Small-to-medium sized  Language Models (MSLMs) capable of integrating visual and textual information for tasks such as image captioning and multimodal question answering. However, MSLMs perform poorly on mathematical problems requiring visual comprehension, including geometric diagrams and spatial reasoning, due to their vision encoders being optimized for general image understanding rather than for capturing the structured, symbolic, and spatial information essential for mathematical tasks. This limits their effectiveness in STEM applications and 
mathematical problem solving.

\textbf{Do MSLMs Truly Comprehend Visual Mathematical Information? } 
The performance gap between text-based and visually presented math problems raises a deeper question: \textit{Do MSLMs actually comprehend mathematical information from visual inputs, or are they simply generating approximations based on weak feature extraction?} While recent MSLMs such as InternLM-XC2 \citep{dong2024internlmxcomposer2masteringfreeformtextimage}, LLaVA-NeXT \citep{liu2024llavanext}, LLaVA-OneVision \citep{li2024llavaonevisioneasyvisualtask}, and Phi-3 \citep{abdin2024phi3technicalreporthighly} excel in general vision-language tasks, including image captioning \citep{vinyals2015tellneuralimagecaption,ge2024visual} and document-based question answering, they struggle with math reasoning tasks that rely on purely visual representations.
\begin{figure*}[!t]
 \centering
\includegraphics[width=1\textwidth]{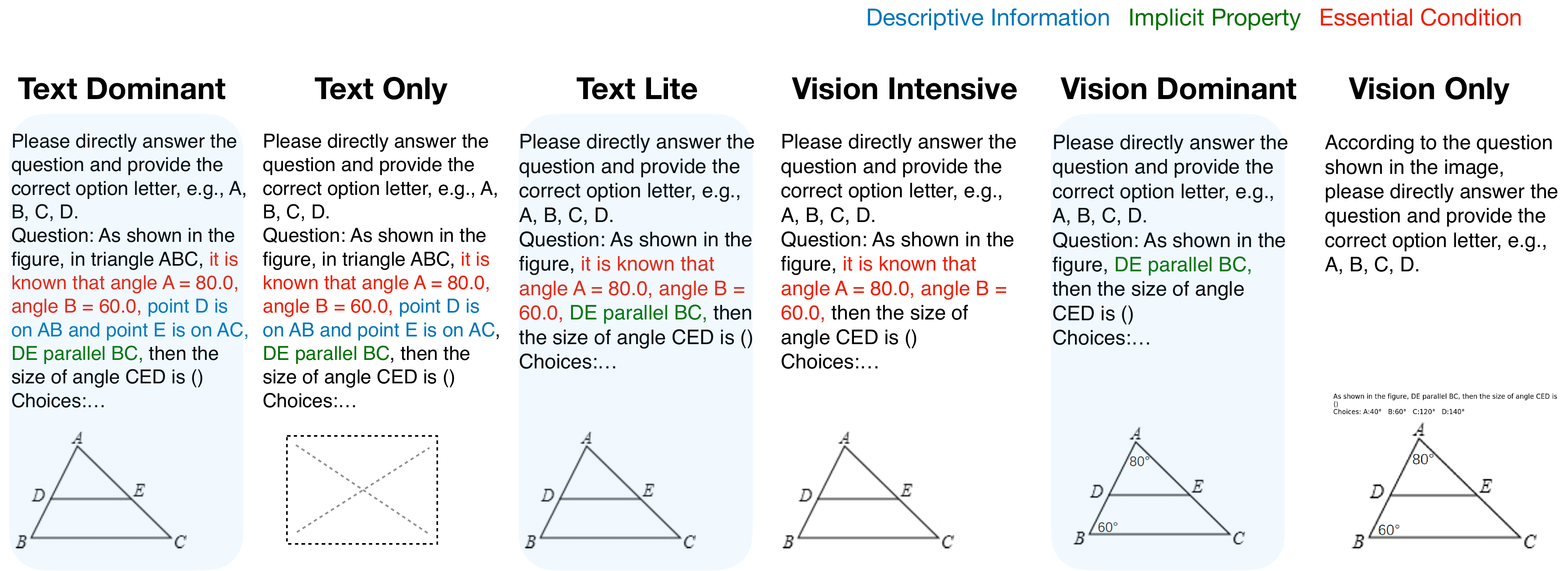}
  \caption{An example from our dataset showcasing six settings of a geometrical mathematical problem.}
  \label{fig:datasample}
  \vspace{-3mm}
\end{figure*}
This limitation is not merely anecdotal -- quantitative evaluations from MathVerse, a recently introduced benchmark for assessing visual mathematical reasoning in MSLMs, highlight these deficiencies \citep{zhang2024mathversedoesmultimodalllm}.
MathVerse evaluates models across varying levels of textual-visual integration (Figure \ref{fig:datasample}), revealing that while MSLMs perform well on text-heavy problems, they fail on vision-dense tasks like diagram-based angle computation and coordinate-based proofs. For example, LLaVA-NeXT-34B achieves 20\% on text-dominant problems but only 9\% on vision-only tasks. Furthermore, another similar study, the DYNAMATH \cite{zou2025dynamathdynamicvisualbenchmark} benchmark, revealed that VLMs consistently produce identical answers despite variations in visual conditions. This behavior suggests a reliance on memorization rather than reasoning grounded in generalized underlying principles. This highlights the need to bridge visual perception with mathematical understanding.

Given the observed deficiencies in MSLMs’ ability to process and reason about visually complex mathematical problems, we investigate the following key research questions:
\begin{itemize}
[itemsep=1pt,topsep=3pt,leftmargin=1em]
    \item \textbf{RQ1:} How effectively do small-to-medium sized multimodal language models (MSLMs) extract and understand structured information from visual mathematical problems?
    \item \textbf{RQ2:} Given an enriched textual spatial representation of a mathematical problem, can small-to-medium sized multimodal language models (MSLMs) generate a structured, stepwise reasoning path leading to the correct solution?
\end{itemize}

To overcome the aforementioned limitations of current MSLMs in mathematical reasoning, we introduce \proposedmodel\ that explicitly infuses visual comprehensions into symbolic reasoning. Our approach consists of:
\begin{itemize}
[itemsep=1pt,topsep=3pt,leftmargin=1em]
    \item \mcmfull. A vision-language model optimized for extracting structured comprehension from visual mathematical content. We formulate \mcm\ that utilizes the structured interpretation generated by an MLLM, also referred as spatial comprehension (SC), during the fine-tuning stage. MSLMs face challenges with generic vision encoders that lack specialization for mathematical content. To address this, we employ Spatial Comprehension Learning, where \mcm\ transforms visual inputs into enhanced textual interpretations, capturing symbolic dependencies and spatial relationships.
    \item \mrmfull. A dedicated solver that integrates spatial comprehension with symbolic reasoning to generate a logically coherent solution. \mrm\ produces step-by-step solutions based on spatial comprehension produced by \mcm. It emphasizes mapping reasoning chains ($r$) for logical consistency and symbolic coherence instead of relying solely on pattern matching. This approach enhances interpretability and accuracy by integrating visual perception with formal mathematical reasoning.
\end{itemize}

We conduct comprehensive experimentation with each module of \proposedmodel.
Notably, \proposedmodel\ achieves a mean accuracy enhancement of 10.2\% and 3.8\% over the standard baseline of supervised fine-tuning with data augmentation on LLaVA-NeXT-34B and Phi-4, respectively. Furthermore, \mcm\ achieves a mean accuracy enhancement of $2.0\%$ (range: 0.8\%-4.8\%) across eight default MSLMs on varying degrees of visual infusion in a problem relative to the default performance.
\begin{figure*}[!t]
\centering
\includegraphics[width=1\textwidth]{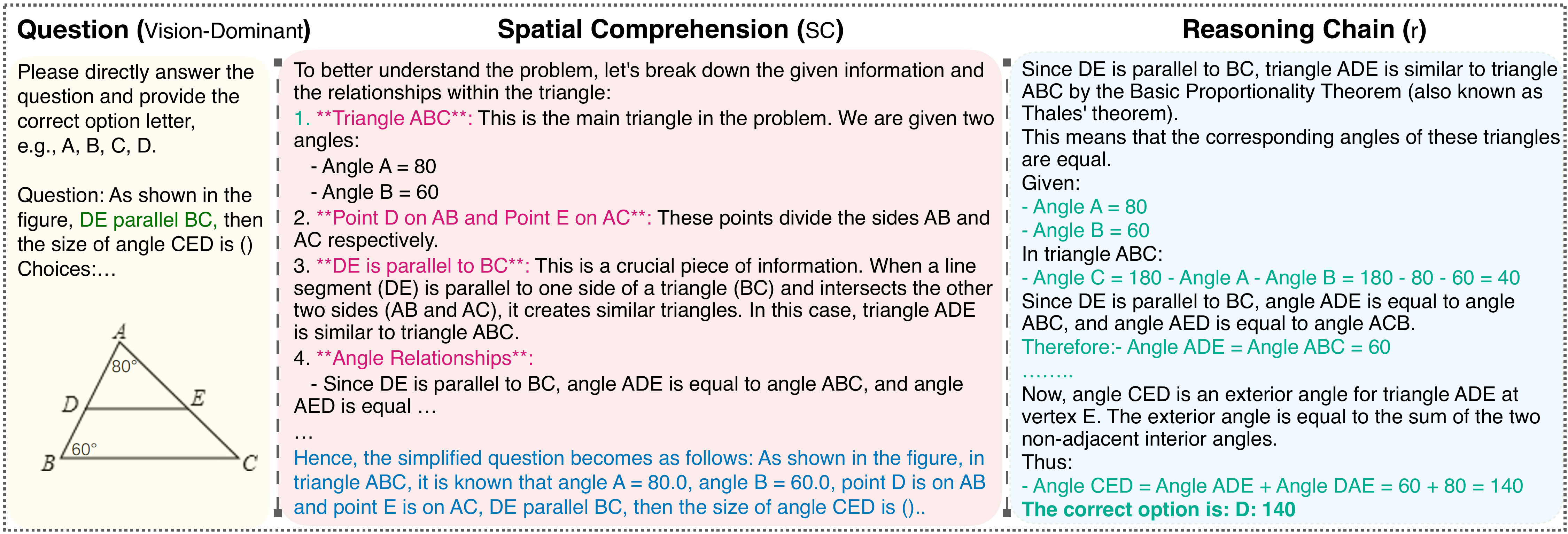}
  \caption{An instance of \spc: the \spcfull\ generated by an MLLM, specifically GPT-4o that includes the text-only variant of the problem highlighted in with blue color, and $r$: the \rcfull, which consists of a series of sequential steps culminating in the final solution, also generated by the aforesaid MLLM.}
  \label{fig:auxdatasample}
  \vspace{-3mm}
\end{figure*}
Our contributions are as follows\footnote{Source code and dataset are available at \url{https://github.com/ab-iitd/spatial-math}}:
\begin{itemize}
[itemsep=1pt,topsep=3pt,leftmargin=1em]
    \item Our research shows that MSLMs struggle significantly with understanding and reasoning through visually intricate mathematical problems, resulting in less effective problem-solving.
    \item To address these challenges, we propose \proposedmodelfull\ (\proposedmodel), a novel methodology to enhance both visual comprehension and symbolic reasoning, supported by \newdataset, a novel dataset that offers spatial comprehensions of complex visual math problems and structured reasoning paths across diverse 
    visual complexity. 
    \item Multi-faceted empirical evaluations demonstrate that \proposedmodel\ consistently outperforms existing baselines, improving both comprehension and problem-solving accuracy in visual mathematical reasoning across datasets and MSLMs.
\end{itemize}

\section{Dataset}
\label{sec:datasetdescription}
We introduce \newdataset, an enhancement of the MathVerse dataset, designed to facilitate our \proposedmodel. MathVerse is the only publicly available dataset that systematically integrates varying levels of visual and textual information into mathematical problems, making it a suitable foundation for evaluating MSLMs’ multimodal reasoning capabilities. Originally derived from multiple public sources \citep{chen-etal-2021-geoqa,lu-etal-2021-inter}, MathVerse reformulates mathematical problems into six different representations -- ranging from text-only to vision-only -- capturing different degrees of visual infusion (see Figure~\ref{fig:datasample}). Building upon this foundation, \newdataset\ extends MathVerse by enriching problem variants with additional structured descriptions and stepwise reasoning annotations as depicted in Figure \ref{fig:auxdatasample}. This augmentation ensures that MSLMs can learn both visual comprehension and symbolic reasoning more effectively. Data construction steps are presented in Algorithm \ref{algo:data_construction}. 

\textbf{Statistics and Quality of \newdataset.}
In the construction of \newdataset, this study focuses exclusively on 2D and 3D geometric problems, deliberately excluding issues related to functions to ensure the dataset is homogenized for geometry-related purposes. Additionally, we consider only five visual modalities derived from the source dataset: Text-Dominant, Text-Lite, Vision-Intensive, Vision-Dominant, and Vision-Only. The Text-Only modality has been excluded, as it does not align with the multimodal focus of this research. \newdataset\ consists of a total of 2.76K training instances, uniformly distributed across all five settings. The test set comprises 500 uniformly distributed instances, reserved for evaluating model performance and reporting results.
\begin{algorithm}[!t]
\small
\caption{Data Construction Approach}  
\begin{algorithmic}
\State \textbf{Input:} Dataset $D$ containing $M$ pairs of geometry questions and corresponding answers.
\State \textbf{Step 1: Define Problem Variants.}
\begin{itemize}
[itemsep=2pt,nolistsep,topsep=0pt,leftmargin=1em]
\item Define a set $K$ consisting of six distinct problem variants, each with different levels of textual and visual integration.  
\end{itemize}
\State \textbf{Step 2: Generate Ground-Truth Spatial Comprehension.}  
\begin{itemize}
[itemsep=2pt,nolistsep,topsep=0pt,leftmargin=1em]
\item Introduce \spcfull\ (\spc), a elaborative structured textual description of the problem's visual elements generated by an LLM serving as the ground-truth representation of the problem's visual content in text-only form.
\item Enhance the dataset with richer linguistic interpretations. 
\end{itemize} 
\State \textbf{Step 3: Enrich the Spatial Comprehension.}  
\begin{itemize}
[itemsep=2pt,nolistsep,topsep=0pt,leftmargin=1em]
\item Enrich the LLM's generated \spcfull\ (\spc) with the text-only form of the problem. 
\item Ensure factual summary of text-based interpretation of visual elements.
\end{itemize} 
\State \textbf{Step 4: Generate Stepwise Reasoning Paths.}  
\begin{itemize}
[itemsep=2pt,nolistsep,topsep=0pt,leftmargin=1em]
\item For each problem variant, leverage a larger model to generate structured \rcfull\ (\rcn), capturing stepwise logical dependencies and symbolic transformations.  
\item Use these structured reasoning paths to derive the correct solution and improve interpretability.  
\end{itemize}   
\State \textbf{Output:} A dataset enriched with textual representations, stepwise reasoning paths, and diverse problem variants.  
\end{algorithmic}
\label{algo:data_construction}
\end{algorithm}

We conduct a manual evaluation of randomly-selected 30 instances. Using a Likert scale \citep{likert} (0–5, with 5 indicating maximum correctness and coherence), we obtain mean scores of $4.36\pm0.86$ and $4.46\pm1.49$ for MLLM generated \spcfulls\ (\spc) and \rcfull\ (\rcn), respectively, demonstrating high reliability of the generated annotations (see Appendix \ref{App:data} for more details on \newdataset).
\begin{figure*}[!t]
 \centering
\includegraphics[width=1\textwidth]{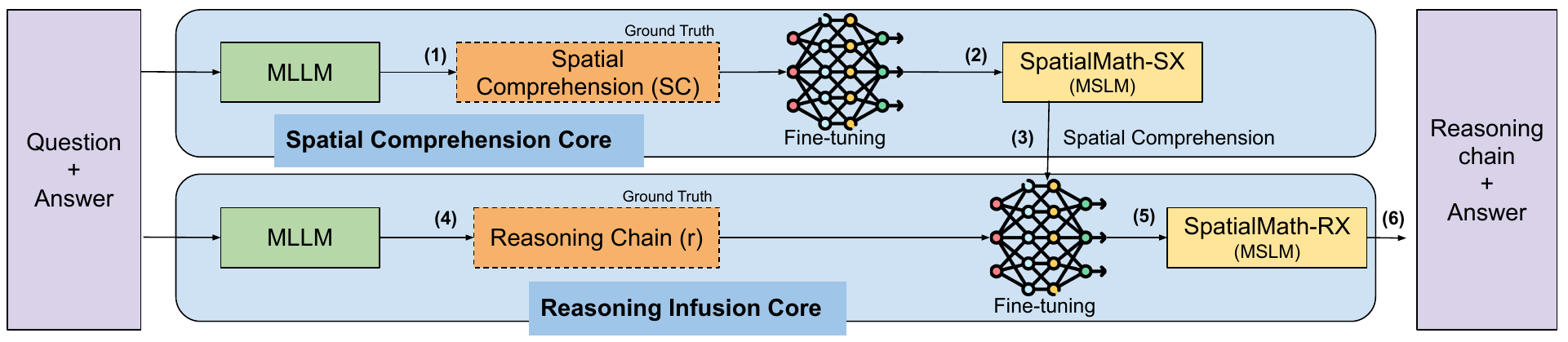}
  \caption{An overview of our framework: (1) Generating \spcfull\ (\spc) for training set using an MLLM; (2) Fine-tuning \mcm\ to enhance visual comprehension; (3) Using \mcm\ at inference to generate \spc\ for test set; (4) Generating \rcfull\ data for train set; (5) Fine-tuning \mrm\ using generated \spc\ for training set by \mcm\ and \rcfull; (6) Deploying \mrm\ at inference for problem-solving.}
  \label{fig:architecture5}
  \vspace{-2mm}
\end{figure*}
\section{Proposed Methodology} 
\label{sec:methodology}
\subsection{Preliminaries}  
We define a dataset \( D \) consisting of \( M \) geometry questions and corresponding answers, represented as pairs \( (q^j_{(t_{(d,i,e)},v)}, a^j) \), where \( j \in \{1, 2, \dots, |M|\} \). Each question \( q^j \) contains textual (\( t \)) and visual (\( v \)) components. The textual component \( t \) is structured into: 
\begin{itemize}[noitemsep,nolistsep,topsep=2pt,leftmargin=1em]
    \item \textbf{Descriptive Information (\( d \))}: Explicitly provided details, i.e., geometric labels and numerical values.
    \item \textbf{Implicit Properties (\( i \))}: Inferred relationships, including parallelism and symmetry.
    \item \textbf{Essential Conditions (\( e \))}: Key algebraic constraints necessary for solving the problem.
\end{itemize}  

The objective is to predict \( a^j \) given the multimodal input \( q^j_{(t_{(d,i,e)},v)} \), formulated as:

\begin{equation}
   a^j  = \arg\max_{a \in A} p\,(a \,| \ q^j_{(t_{(d,i,e)},v)})
   \label{eq:maxlikelihood}
\end{equation}  

where \( A \) is the answer space. 

\subsubsection{Problem Variants and Transformations} 
We extend the MathVerse formulation with varying textual and visual integration to five settings and exclude the Text-Only modality to align with the multimodal focus of this research. 
\begin{itemize}
[noitemsep,nolistsep,topsep=2pt,leftmargin=1em]
\small
    \item \textbf{Text-Dominant} (\( q^j_{(t_{(d,i,e)},v)} \)): All textual and visual input.
    \item \textbf{Text-Lite} (\( q^j_{(t_{(i,e)},v)} \)): Text lacks descriptive details (\( d \)), requiring reliance on \( i \) and \( e \).
    \item \textbf{Vision-Intensive} (\( q^j_{(t_{(e)},v)} \)): Only essential conditions (\( e \)) are retained in text; implicit properties (\( i \)) must be inferred from the visual input.
    \item \textbf{Vision-Dominant} (\( q^j_{(t_{(i)},v^{'})} \)): Only implicit properties (\( i \)) in Text, while numerical values are visually embedded.
    \item \textbf{Vision-Only} (\( q^j_{(_,v^{''})} \)): The problem is presented purely visually, with no textual input.
\end{itemize}  
Here, \( v' \) enriches the diagram with essential numeric labels, while \( v^{''} \) is fully self-contained, requiring complete visual parsing. 

\subsection{\mcm}  
While MSLMs aim to jointly comprehend and solve visually complex mathematical problems, their visual comprehension capabilities remain a bottleneck. To address this, we introduce \mcmfull, a dedicated module that independently evaluates and enhances the model's ability to interpret visual information before reasoning over it.

\subsubsection{Spatial Comprehension}  
\mcm\ is trained on an auxiliary dataset \( D' \), derived from \( D \), where each question-answer pair \( (q^j, a^j) \) in \( D \) is transformed into a question-comprehension pair \( (q^k, c^{sc}) \) in \( D' \). Here, \( c^{sc} \) represents spatial comprehension of the question, extracted from an MLLM that explicitly describes key visual and textual elements. The transformation allows \mcm\ to focus solely on the comprehension, independent of reasoning.  

The objective of \mcm\ is to maximize the likelihood of generating the most accurate spatial comprehension \spc\ given the multimodal question \( q^k \), is presented in Equation \ref{eq:maxlikelihood2}, where \( B \) denotes the vocabulary space of possible comprehensions. Additionally, to enrich the extracted comprehension, we integrate text-only variant of the problem with LLM generated \spc, resulting in the enhanced objective:
\begin{equation}
   \spc^k_{t_{(d,i,e)}}  = \arg\max_{sc \in B} p\,(sc\,| \ q^k_{(t^*,v^*)})
   \label{eq:maxlikelihood2}
\end{equation}   
\subsubsection{Learning Objective}  
Many mid-sized multimodal models struggle with effective visual comprehension. To improve this, we fine-tune a dedicated \mcm\ using parameter-efficient fine-tuning (PEFT) \citep{han2024parameterefficientfinetuninglargemodels,hu2021loralowrankadaptationlarge}. Specifically, we apply Low-Rank Adaptation (LoRA) to the multi-layer perceptron (MLP) projector within the pre-trained LLaVA-NeXT-34B model while keeping the LLM and vision encoder frozen. A similar setup is followed for Phi-4.  

\mcm\ is trained by minimizing the negative log-likelihood (NLL) loss across all training instances \( n' \):
\vspace{-2mm}
\begin{equation}
\mathcal{L}_{SX} = - \sum_{k=1}^{n'} \log P(\spc^{k}_{t_{(d,i,e)}} | \, q^k_{(t^*,v^*)})     
   \label{eq:loss}
\end{equation}
\vspace{-3mm}
\subsection{\mrm}
An \mcm\ can interpret the visual aspects of math problems and bridge different problem formats in vision-only settings but does not improve step-wise reasoning toward the correct solution. Mid-sized models often struggle with multi-step deduction, and multimodal objectives rarely align spatial features with structured reasoning. To address this, we propose \mrmfull, which embeds spatial comprehension into reasoning chains for context-aware, step-wise symbolic reasoning.
\begin{table*}[t!]
\centering
\small
\resizebox{1\textwidth}{!}{
\begin{tabular}{lcccccccccccc}
\toprule
\textbf{Models}& \multicolumn{2}{c}{\textbf{All}} &\multicolumn{2}{c}{\textbf{Text Dominant}} &\multicolumn{2}{c}{\textbf{Text Lite}}& \multicolumn{2}{c}{\textbf{Vision Intensive}}& \multicolumn{2}{c}{\textbf{Vision Dominant}}&  \multicolumn{2}{c}{\textbf{Vision Only}}\\
\midrule
\multicolumn{13}{l}{\textbf{(a): LLaVA-NeXT-34B as Base Model for both} \mcm\ \textbf{and} \mrm} \\
\multicolumn{13}{l}{\textbf{(b): Phi-4 as Base Model for both} \mcm\ \textbf{and} \mrm}\\
\midrule
& \textbf{(a)}& \textbf{(b)} & \textbf{(a)}& \textbf{(b)} & \textbf{(a)}& \textbf{(b)} &\textbf{(a)}& \textbf{(b)} &\textbf{(a)}& \textbf{(b)} &\textbf{(a)}& \textbf{(b)} \\
\midrule
Zero-Shot&18.0&28.2&25.0&	34.0&18.0&	30.0&21.0&	30.0&13.0&	27.0&13.0&	20.0\\
ICL (1-shot) &10.0&28.4	&15.0&	32.0&	10.0&	33.0	&5.0&	31.0&	11.0&	22.0	&9.0&	24.0\\
CoT (1-shot) &17.2&28.0&	27.0&	 31.0&	18.0&	33.0&	15.0&	30.0&	17.0&	22.0&	9.0&	24.0\\
SFT &18.0&33.4&22.0&	38.0&20.0&	35.0&22.0&	33.0&15.0&	30.0&11.0&	31.0\\
SFT +  Data Augmentation&19.2&24.6&21.0&	35.0&23.0&	25.0&18.0&	22.0&\textbf{20.0}&	22.0&14.0&	19.0\\
\midrule
\proposedmodel&\textbf{23.0}&\textbf{43.6}&27.0&	\textbf{53.0}&\textbf{23.0}&	\textbf{48.0}&\textbf{28.0}&	\textbf{45.0}&17.0&	\textbf{41.0}&\textbf{20.0}&\textbf{31.0}\\
\bottomrule
\end{tabular}
}
\caption{Comparison of \proposedmodel\ with baselines across various problem settings on accuracy scores. one tail Mann-Whitney U test for p-values. We observe one tail Mann-Whitney U test-based p-value of 0.03 and 0.07 for Phi-4 and LLaVA-based \proposedmodel\ when compared to nearest baselines across visual infusions, respectively. }
\label{tab:mainres}
\end{table*}
\begin{table*}[t!]
\centering
\small
\resizebox{1\textwidth}{!}{
\begin{tabular}{lcccccccccccc}
\toprule
\textbf{Solver Models}& \multicolumn{2}{c}{\textbf{All}} &\multicolumn{2}{c}{\textbf{Text Dominant}} &\multicolumn{2}{c}{\textbf{Text Lite}}& \multicolumn{2}{c}{\textbf{Vision Intensive}}& \multicolumn{2}{c}{\textbf{Vision Dominant}}&  \multicolumn{2}{c}{\textbf{Vision Only}}\\
\midrule
\multicolumn{13}{l}{\textbf{(a): Default Solver Performance (Zero-shot Prompting)}} \\
\multicolumn{13}{l}{\textbf{(b): Default Solver Performance with LLaVA-NeXT-34B-based} \mcm}\\
\midrule
& \textbf{(a)}& \textbf{(b)} & \textbf{(a)}& \textbf{(b)} & \textbf{(a)}& \textbf{(b)} &\textbf{(a)}& \textbf{(b)} &\textbf{(a)}& \textbf{(b)} &\textbf{(a)}& \textbf{(b)} \\
\midrule
{Phi3.5}& 22.4& \cellcolor{LighterGreen}{25.4}&	23.0&	\cellcolor{LighterGreen}{25.0}&	20.0&\cellcolor{LighterGreen}{30.0}&	23.0&	\cellcolor{LighterGreen}{26.0}	&26.0&22.0&	20.0&\cellcolor{LighterGreen}{24.0}\\
{InternLMXC2}& 22.4& \cellcolor{LighterGreen}{23.6}	&25.0&23.0	&26.0&	\cellcolor{LighterGreen}{27.0}	&29.0&25.0	&15.0&\cellcolor{LighterGreen}{26.0}	&17.0 &17.0\\
{LLaVA-NeXT-34B} &18.0&\cellcolor{LighterGreen}{19.4}&	25.0&\cellcolor{LighterGreen}{28.0}&	18.0&	18.0&	21.0&	19.0&	13.0&	\cellcolor{LighterGreen}{15.0}&	13.0&	\cellcolor{LighterGreen}{17.0}\\
{LLaVA OV 7B}&30.6&\cellcolor{LighterGreen}{35.4}	&34.0&\cellcolor{LighterGreen}{43.0}	&36.0&\cellcolor{LighterGreen}{38.0}	&34.0&	\cellcolor{LighterGreen}{38.0}	&26.0&	\cellcolor{LighterGreen}{33.0}	&23.0&\cellcolor{LighterGreen}{25.0}\\
{LLaVA OV 70B}&34.4	&\cellcolor{LighterGreen}{35.4}&46.0&	\cellcolor{LighterGreen}{50.0}	&36.0&	\cellcolor{LighterGreen}{39.0}&	37.0&	36.0	&34.0&	33.0&	19.0&	19.0\\
MultiMath-7B $\color{blue}{\diamond }$&24.4&\cellcolor{LighterGreen}{25.8}&33.0&\cellcolor{LighterGreen}{36.0}&28.0&23.0&23.0&\cellcolor{LighterGreen}{29.0}&29.0&26.0&9.0&\cellcolor{LighterGreen}{15.0}\\

Math-LLaVA $\color{blue}{\diamond }$&21.6&\cellcolor{LighterGreen}{23.4}	&27.0&	22.0&	21.0&	\cellcolor{LighterGreen}{27.0}&	25.0&\cellcolor{LighterGreen}{32.0}&16.0&\cellcolor{LighterGreen}{21.0}&	19.0&15.0\\

G-LLaVA $\color{blue}{*}$&22.6&\cellcolor{LighterGreen}{23.4}&	23.0&22.0	&	23.0	&\cellcolor{LighterGreen}{26.0}&22.0&\cellcolor{LighterGreen}{26.0}&	23.0&	20.0&	22.0&\cellcolor{LighterGreen}{23.0}\\

\bottomrule
\end{tabular}
}
\caption{Performance increase across various MSLMs attributed to \spcfull\ produced by the LLaVA-NeXT-34B-based \mcm\ ($\color{blue}{\diamond }$ and $\color{blue}
{*}$ represent Math- and domain-focused MSLMs, respectively).}
\label{tab:sc_generalization}
\vspace{-3mm}
\end{table*}
\subsubsection{Learning Objective} To this end, we leverage the \rcfull\ (\rcn) generated by a comparatively larger model to enhance the performance of a medium-sized solver. The solver model is designed to first generate a \rcfull, $r^k$, then derive the final answer $a^k$ given $q^k_{(t^*,v^*)}$ and \spc, ${(\spc}^k_{t_{(d,i,e)}}$. Unlike \mcm, which is trained using structured spatial comprehension labels (\spc), \mrm\ is fine-tuned using a dataset explicitly enriched with structured reasoning steps (\rcn) to facilitate progressive mathematical inference. 
\vspace{-2mm}
\begin{equation}
\mathcal{L}_{RX} = - \sum_{k=1}^{n'} \log P(r^k, a^k \,| \,q^k_{(t^*,v^*)},\spc^k_{t_{(d,i,e)}}  )     
   \label{eq:loss_solver}
\end{equation}
In the process, an explicit alignment is achieved between spatial comprehension and the structured symbolic reasoning steps. The enhanced comprehension is leveraged in the reasoning infusion core (\mrm), which builds upon \mcm's extracted information to perform accurate problem-solving. We employ the LLaVA-NeXT-34B and Phi-4 models as the base models, utilizing a similar PEFT strategy as \mcm. The joint optimization for reasoning follows in Equation \ref{eq:loss_solver}. 

Finally, the summarized formulation \proposedmodel\ is composed of both the spatial comprehension core and the reasoning infusion core (\mcm\ + \mrm) as shown in Figure \ref{fig:architecture5}:

\section{Experimental Setup}
\label{sec:exp_setup}
We conduct rigorous experiments using LLaVA- and Phi-based architectures to evaluate the proposed method. 
The dataset is split into 80\% train and 20\% test sets. We use GPT-4o to generate comprehension cues and reasoning chains for data augmentation. All hyperparameters, prompts, and other details are provided in the Appendix \ref{App:hyperparameter} and Appendix \ref{App:prompts}, respectively.
\paragraph{Evaluation Metrics.}
We use accuracy as the primary metric, assigning 1 for correct and 0 for incorrect responses. Additionally, we employ Gemini 1.5 Pro \citep{geminiteam2024gemini15unlockingmultimodal} for post-processing model outputs to extract \& match the final answer.
\paragraph{Models.}  
LLaVA-NeXT-34B and Phi-4 are two primary backbones, used for modling \proposedmodel. To evaluate generalization, we experiment with general, math-, and geometry-focused MSLMs of various sizes;
{LLaVA-NeXT-7B, LLaVA-OV-7B, InternLM-XC2, Phi-4 \citep{abdin2024phi4technicalreport}, MultiMath-7B \citep{peng2024multimathbridgingvisualmathematical}, MATH-LLaVA, G-LLaVA \citep{gao2023gllavasolvinggeometricproblem}, and LLaVA-OV-70B \citep{li2024llavaonevisioneasyvisualtask}}.

\paragraph{Baselines.}  
We compare \proposedmodel\ against five strong baselines: (1) Zero-shot: direct model response without additional training.  (2) In-context learning (ICL) \citep{brown2020languagemodelsfewshotlearners}: providing 3-shot demonstrations in the prompt. (3) Chain-of-thought (CoT) prompting \citep{wei2023chainofthoughtpromptingelicitsreasoning}: prompting the model to explicitly generate intermediate reasoning steps before answering. (4) Supervised fine-tuning (SFT) \citep{radford2021learningtransferablevisualmodels}: fine-tuning directly on ($q^j, a^j$) pairs from the training set.  (5) SFT + Data Augmentation: using \spcfull\ (\spc) and \rcfull\ ($r$) to fine-tune a standalone solver, inspired by MultiMath-7B \citep{peng2024multimathbridgingvisualmathematical}.

\section{Results}
\label{sec:results}
\subsection{Performance Comparison}
\proposedmodel\ demonstrates substantial performance gains over baseline methods, particularly in vision-dense settings across both base models--LLaVA-NeXT-34B and Phi-4. As shown in Table \ref{tab:mainres}, for LLaVA-NeXT-34B, this approach yields a 10.0 percentage point improvement in vision-intensive settings and a 6.0 percentage point improvement in vision-only settings compared to the Supervised Fine-tuning (SFT) + Data Augmentation baseline. When averaged across all settings, \proposedmodel\ achieves a 3.8 percentage point mean accuracy improvement, demonstrating robustness across formats. For Phi-4, it outperforms SFT baseline with a notable 10.2 percentage point mean accuracy gain across all settings. These findings highlight the strength of infusing \spcfull\ into structured \rcfull\ for addressing visually complex mathematical problems.
\begin{figure}[!t]
\centering
\includegraphics[width=1\columnwidth]{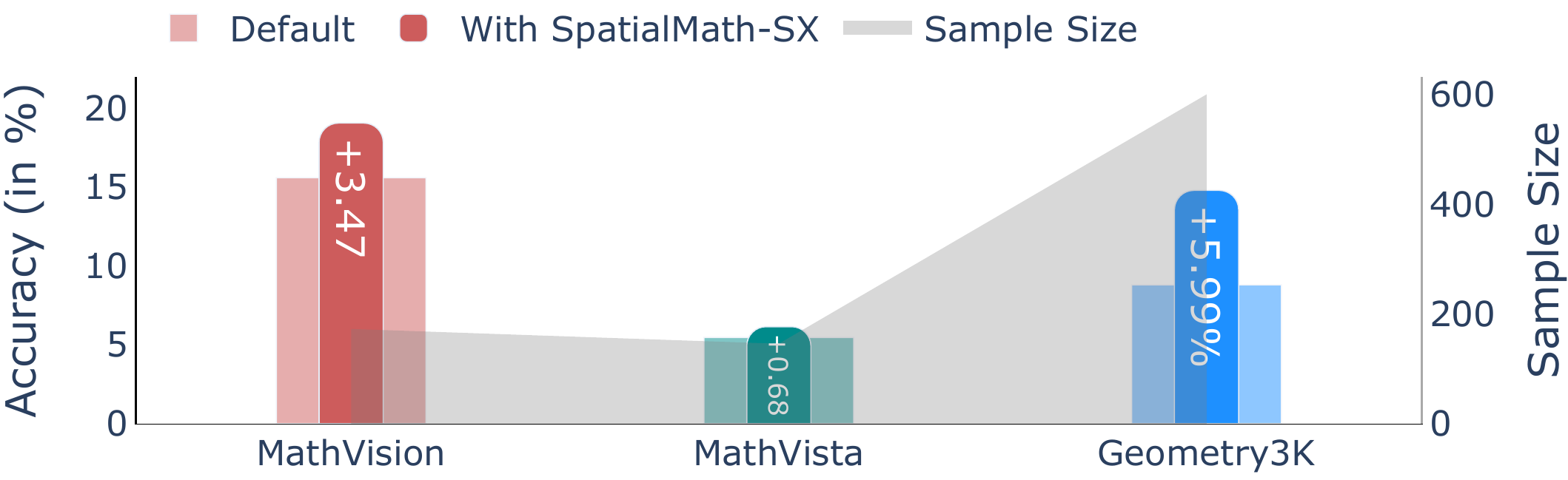}
  \caption{Performance of LLaVA-NeXT-34B-based \mcm\ alogn with default LLaVA-NeXT-34B-based solver across three other public benchmarks- MathVision, MathVista, and Geometry3K.}
  \label{fig:benchmarkdata}
  \vspace{-3mm}
\end{figure}
\subsection{Robustness: \mcm}
\subsubsection{The significance of Spatial Comprehension Core Across MSLMs}
As shown in Table \ref{tab:sc_generalization}, passing \spcfull\ extracted by \mcm\ as an additional context to default MSLMs yields consistent improvements across various settings. Specifically, we observe accuracy gains up to 7.0, 11.0, and 6.0 percentage points for vision-only, vision-dominant, and vision-intensive settings, respectively, leading to an overall mean accuracy improvement of 2.0 (range: 0.8-4.8) percentage points across all settings and solver models. Notably, in the vision-dominant setting, InternLM-XC2 and LLaVA-OV-7B achieve substantial gains of 11.0 and 7.0 percentage points, respectively, while LLaVA-NeXT-34B exhibits a 4.0 percentage point boost in the vision-only setting. These results highlight the robustness and scalability of \mcm’s structured multimodal comprehension, particularly when applied to specialized visual mathematical models such as MultiMath-7B, reinforcing its efficacy in diverse multimodal reasoning tasks. In response to RQ1, the empirical results indicate that the performance of MSLMs declines as the complexity of visual inputs increases, struggling to understand and process intricate visual information effectively. Further, findings indicate that MSLMs with enriched textual spatial representation and reasoning chains improve visual math problem-solving in response to RQ2.
\begin{table}[t!]
\centering
\small
\resizebox{\columnwidth}{!}{
\begin{tabular}{lccccc}
\toprule
\textbf{Setting} &\textbf{BLEU4}&\textbf{R1} &\textbf{RLSum} &\textbf{Met} & \textbf{BERTScore}\\
\midrule
\multicolumn{6}{l}{\textbf{\mcm:} LLaVA-NeXT-34B}\\
Default Zero-shot & 8.8	 &49.0&	45.6&	22.9&	84.7\\
\mcm &\textbf{38.3}	&\textbf{66.8}	&\textbf{64.7}&	\textbf{54.2}&	\textbf{89.7}\\
\midrule
\multicolumn{6}{l}{\textbf{\mcm:} LLaVA-NeXT-7B}\\
Default Zero-shot & 11.4 & 47.9 & 44.9 & 25.8 & 84.4 \\

 \mcm &\textbf{14.5} &\textbf{51.0} &\textbf{49.2} &\textbf{41.3} &\textbf{87.2}\\

\bottomrule
\end{tabular}
}
\caption{Implicitly evaluate the fine-tuned \mcm\ by comparing generated \spc\ with ground truth using lexical and semantic metrics.}
\label{tab:sc_evaluation}
\vspace{-3mm}
\end{table}
\subsubsection{Evaluation Across Public Benchmarks}
To evaluate data generalization, we used three recent public visual-math benchmarks—MathVision (173 samples), MathVista (146), and Geometry3K (601)—which include out-of-distribution samples. To mitigate out-of-domain issues, we focus exclusively on geometry-based problems within these datasets. Figure \ref{fig:benchmarkdata} illustrates the performance difference of default MSLM's with and without the \mcm's output as additional context, utilizing the LLaVA-NeXT-34B-based fine-tuned model trained on MathVerse training data. The results demonstrate that MSLM's with \spcfull\ yields improvements of 3.47, 0.68, and 5.99 percentage points for MathVision, MathVista, and Geometry3K, respectively, thereby corroborating the robust generalization across other datasets. 

\subsubsection{Implicit Evaluation}
We evaluate fine-tuning robustness by assessing \spc\ quality across \mcm\ models of different sizes, beyond task accuracy. Generated \spc\ are compared to GPT-4o ground-truth using BLEU, ROUGE (R1, RLSum), METEOR, and BERTScore. Table \ref{tab:sc_evaluation} shows fine-tuned \mcm\ consistently outperforming zero-shot models. Mean improvements of 16.3 (BLEU-4), 10.45 (R1), 11.7 (RLSum), 23.4 (METEOR), and 3.9 (BERTScore) confirm fine-tuning enhances \spc\ richness and alignment with ground truth.
\begin{table*}[t!]
\centering
\small
\resizebox{\textwidth}{!}{
\begin{tabular}{lcccccccccccc}
\toprule
\textbf{Solver Models}& \multicolumn{2}{c}{\textbf{All}} &\multicolumn{2}{c}{\textbf{Text Dominant}} &\multicolumn{2}{c}{\textbf{Text Lite}}& \multicolumn{2}{c}{\textbf{Vision Intensive}}& \multicolumn{2}{c}{\textbf{Vision Dominant}}&  \multicolumn{2}{c}{\textbf{Vision Only}}\\
\midrule
\multicolumn{13}{l}{\textbf{\mrm: LLaVA-NeXT-7B}}\\
\multicolumn{13}{l}{\mcm: (a): LLaVA-NeXT-34B  (b): LLaVA-NeXT-7B}\\
\midrule
& \textbf{(a)}& \textbf{(b)} & \textbf{(a)}& \textbf{(b)} & \textbf{(a)}& \textbf{(b)} &\textbf{(a)}& \textbf{(b)} &\textbf{(a)}& \textbf{(b)} &\textbf{(a)}& \textbf{(b)} \\
\midrule

Zero-shot&18.0&14.2	&25.0&	17.0	&18.0&	13.0	&21.0&15.0	&13.0&15.0	&13.0&11.0\\
\mcm\ + default solver&12.4&15.2	&12.0&	13.0		&11.0&17.0	&15.0&17.0	&13.0&15.0	&11.0&	14.0\\
\proposedmodel &15.4&13.6	&14.0&	14.0	&13.0&13.0	&18.0&14.0		&18.0&13.0	&14.0&	14.0\\
\midrule

\multicolumn{13}{l}{\textbf{\mrm: LLaVA-OV-7B}} \\
\multicolumn{13}{l}{\mcm: (a): LLaVA-NeXT-34B  (b): LLaVA-OV-7B}\\

\midrule

Zero-shot&30.6&30.6&	34.0&	34.0&	36.0&	36.0&	34.0&	34.0&	26.0&	26.0&	23.0&	23.0\\
\mcm\ + default solver&35.6&30.2&	44.0&	39.0&	38.0&	32.0&	38.0&	30.0&	33.0&	28.0&	25.0&	22.0\\
\proposedmodel&30.0&22.6&	40.0&	26.0&	32.0&	24.0&	28.0&	24.0&	25.0&	17.0&	25.0&	22.0\\
\bottomrule
\end{tabular}
}
\caption{Comparison of two distinct \mcm\ models, focusing on a diverse range of parameters associated with two \mrm\ -- LLaVA-NeXT-7B and LLaVA-OV-7B, across a variety of problem settings.}
\label{tab:solver_gen}
\vspace{-3mm}
\end{table*}
\subsection{Generalization: \mrm}
To assess \mrm's generalization, we examine if a reasoning solver improves with \spcfull\ from different specialized \mcm\ models. Specifically, we investigate two 7B-sized \mrm\ -- LLaVA-NeXT-7B and LLaVA-OV-7B, and assess their performance using \spc\ generated by fine-tuned \mcm\ variants. For LLaVA-NeXT-7B, we consider two \mcm\ variants—LLaVA-NeXT-34B and LLaVA-NeXT-7B, while for LLaVA-OV-7B, we use LLaVA-NeXT-34B and LLaVA-OV-7B-based \mcm.

Table \ref{tab:solver_gen} shows that using \spc\ generated from the larger  \mcm\ (LLaVA-NeXT-34B) model improves LLaVA-OV-7B's mean accuracy by 5.0\% points across all settings. This confirms that even relatively smaller \mrm\ can benefit from multimodal interpretations generated by a more capable \mcm. However, we observe an inconsistent decline in performance when \mrm\ rely on \spc\ from smaller (7B-sized) \mcm. Notably, LLaVA-OV-7B exhibits an 8.0 percentage point drop in accuracy when using \spc\ generated from LLaVA-OV-7B itself, highlighting the limitations of smaller models in producing visual interpretations-- even after fine-tuning. 
\begin{figure}[!t]
\centering
\includegraphics[width=1\columnwidth]{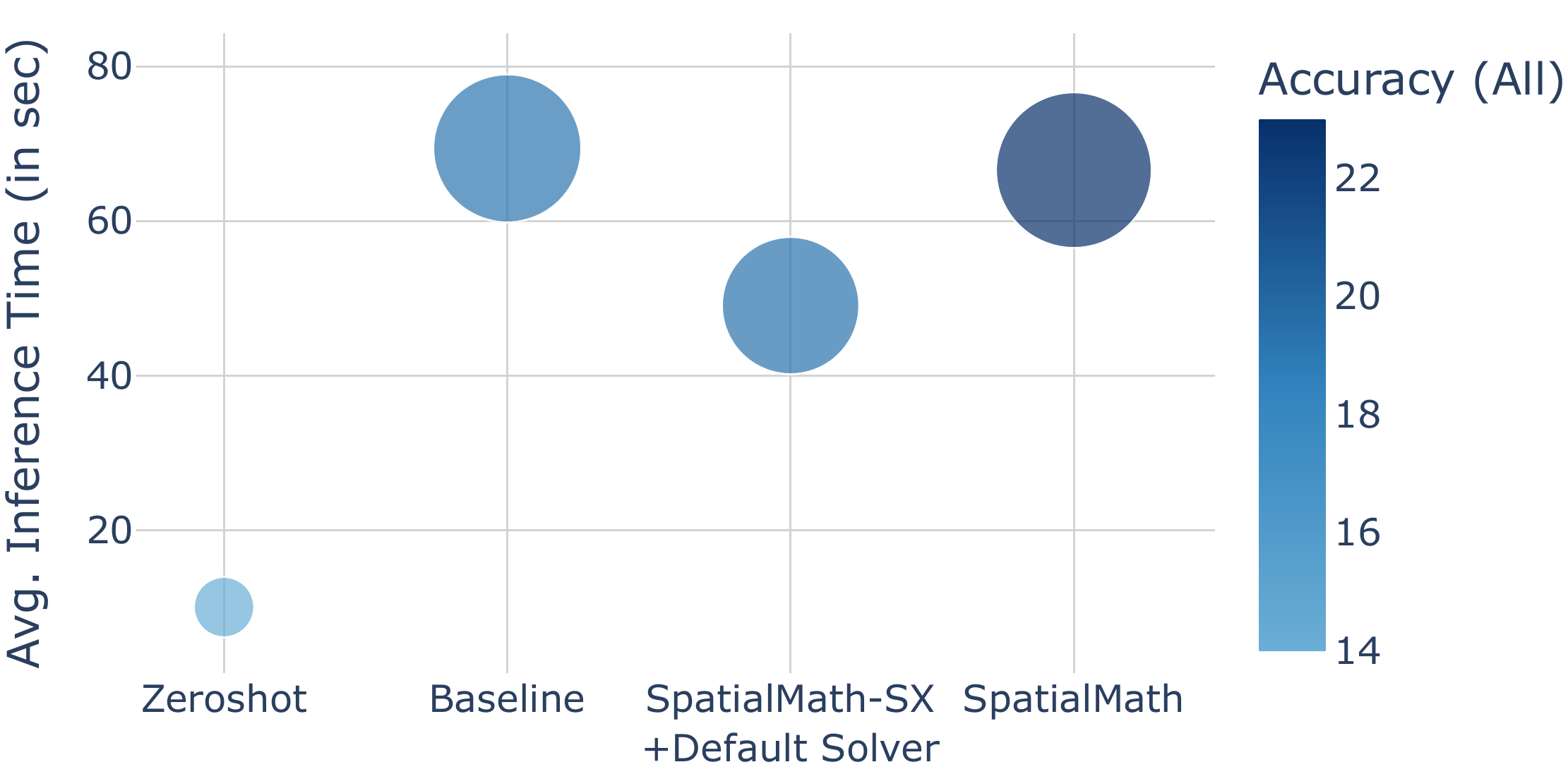}
  \caption{Analysis on computational overhead in comparison to a single model baseline (SFT+Data Aug.) approach. Bubble Size: Average \#tokens generated}
  \label{fig:inf_time}
  \vspace{-3mm}
\end{figure}
\subsection{Evaluator Modeling: \mcm}
 Inadequate quality of the \mcm's output (\spc) may adversely affect the \mrm\ core's performance. To address this, we introduce an intermediate evaluation step for \spc\ between the \mcm\ and \mrm\ cores. To this end, we finetune an evaluative model for a binary decision to determine whether the \spc\ is useful in relation to the underlined question. Based on the decision, we proceed to pass the \spc\ to \mrm\ exclusively when the evaluator model issues a "yes" response. To evaluate the impact of this evaluative model, we utilize the improvement ratio, defined as the overall improvement divided by the overall degradation in utilizing \spc\ with \mrm\, to quantify the effectiveness of quality control. We compare two configurations: \mcm\ with and without the evaluative model across both baseline models, LLaVA-NeXT-34B and Phi-4. Notably, evaluative model enhances the improvement ratio from 1.13 to 2.0 for LLaVA-NeXT-34B and from 1.12 to 3.0 for Phi-4, respectively. (refer to  Appendix \ref{app:evaluator} for more details on evaluator modeling and metric.)

\subsection{Computational Efficiency Analysis}
The experimental results from Figure \ref{fig:inf_time} demonstrate that \proposedmodel\ method is computationally efficient compared to the integrated solver (SFT + data augmentation) approach. While both methods achieve comparable inference time, with \proposedmodel\ outperforming on all metrics (Accuracy: 23 vs. 19.2) and lower average inference time (66.59 vs. 69.41 seconds). This highlights the method's effectiveness in balancing computational overhead with improved performance outcomes. Inference times are averaged over three runs on a random set of 100 samples. 
Additionally, we report the total number of tokens (word-level) generated by the model in each settings. Furthermore, \proposedmodel\ is 16× more computationally efficient, requiring only two A100 80GB GPU compared to MathPUMA’s 32 A100 80GB GPUs.

\subsection{Error Analysis}
\begin{figure}[!t]
\centering
\includegraphics[width=1\columnwidth]{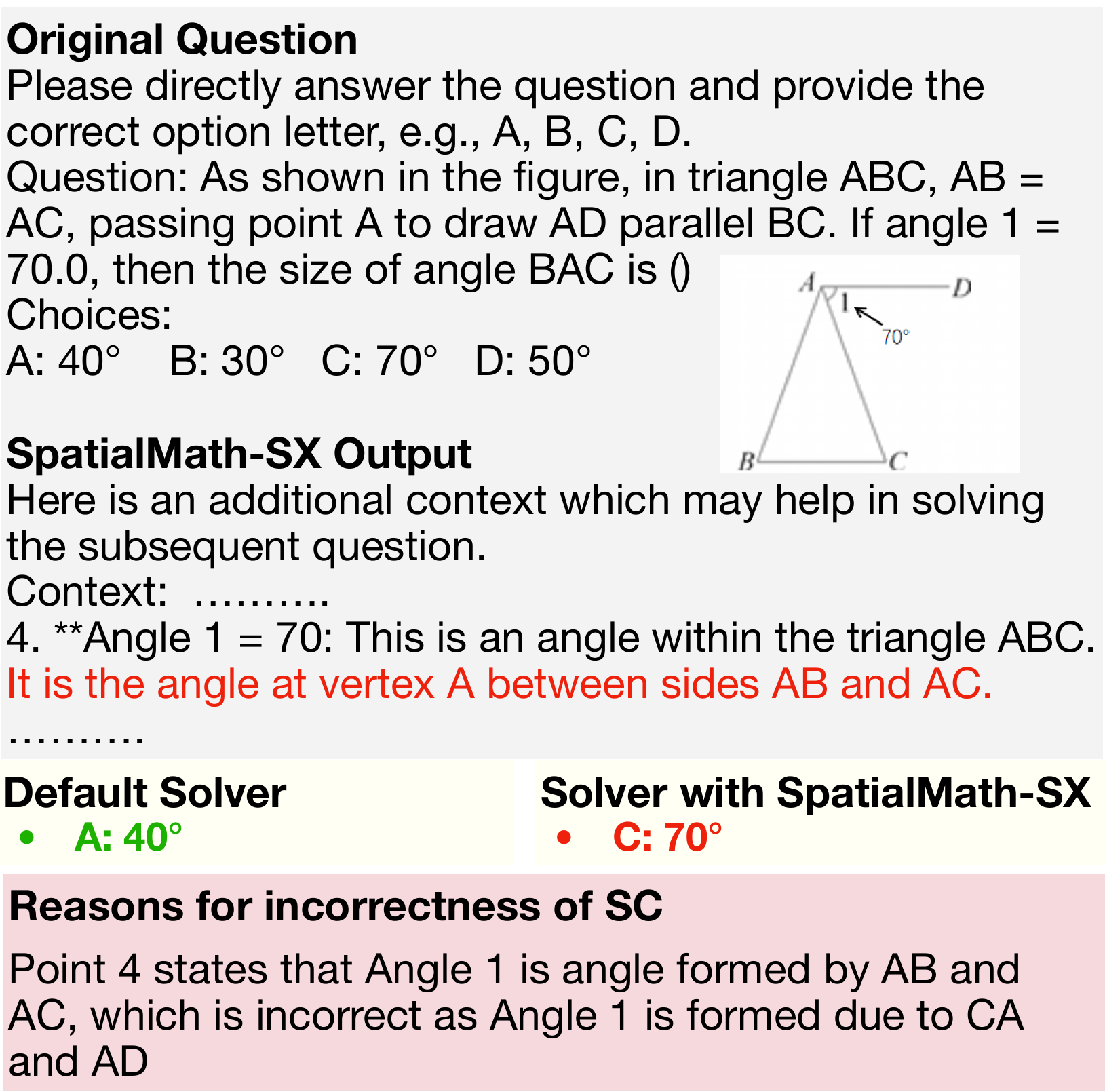}
  \caption{An example of the unfavorable consequences of misleading spatial comprehension caused by the \mcm\ on the \mrm.}
  \label{fig:incorrect_aux}
  \vspace{-3mm}
\end{figure}

Our failure case analysis reveals that a primary source of error stems from inaccurate \spc\ generated by the \mcm,  affecting the solver's response quality adversely.  Figure \ref{fig:incorrect_aux} highlights an example from a text-dominant setting, where an LLaVA-NeXT-34B-based \mcm\ produces an incorrect comprehension of the problem. This erroneous \spc\ leads LLaVA-NeXT-34B-based \mrm\ to deviate from the correct solution, ultimately generating an incorrect answer. These findings emphasize the importance of quality control mechanisms in refining \mcm\ outputs, suggesting potential solutions on error mitigation strategies.  Accordingly, we explore an intermediate step to evaluate the quality of \spcfull, generate by \mcm, prior to forwarding them to the \mrm. In the future, research community can investigate alternative approaches, such as reinforcement learning from human feedback or cross-context validation to address these challenge.
\section{Related Work}
\textbf{Mathematical Reasoning in LLMs.} Advancements in MLLMs, such as GPT-4o \citep{openai2024gpt4technicalreport} and Gemini \citep{geminiteam2024geminifamilyhighlycapable}, have significantly improved context understanding \citep{brown2020languagemodelsfewshotlearners} and logical reasoning \citep{10.5555/3600270.3602070}, enabling applications in mathematical reasoning and word problem-solving. Early research, including MATH \citep{hendrycks2021measuringmathematicalproblemsolving} and MWP-BERT \citep{liang-etal-2022-mwp}, focused on text-only problem-solving, excel in algebraic manipulation and numerical inference but struggle with spatial reasoning \citep{lu-etal-2023-survey}.

\paragraph{Visual Mathematical Reasoning.} 
Models such as LLaVA \citep{liu2023visualinstructiontuning}, InternLM-XC \citep{Zhang2023InternLMXComposerAV}, Phi-3 \citep{abdin2024phi3technicalreporthighly}, and Pixtral \citep{agrawal2024pixtral12b} have shown strong performance in visual understanding tasks such as image captioning and VQA \citep{agrawal2016vqavisualquestionanswering}, demonstrating impressive accuracy \citep{Liu_2024_CVPR} but require additional techniques to align symbolic notations with their visual representations when applied in mathematical contexts \citep{yan2024surveymathematicalreasoningera}.
Dedicated efforts to integrate symbolic reasoning with diagrammatic understanding have led to specialized multimodal models, including MATH-Vision \citep{wang2024measuringmultimodalmathematicalreasoning}, MATHVISTA \citep{lu2024mathvista}, Math-LLaVA \citep{shi-etal-2024-math}, and CLEVR-Math \citep{lindström2022clevrmathdatasetcompositionallanguage}. They aim to preserve the syntactic integrity of mathematical expressions while incorporating spatial reasoning from diagrams.

\paragraph{Novelty of Our Work.} MathVerse \citep{zhang2024mathversedoesmultimodalllm}  analyzes visual mathematical reasoning by assessing model performance with different levels of visual infusion. However, it lacks a systematic method for improving performance in vision-dense contexts, a critical gap that our research effectively addresses. Early approaches, such as MAVIS \citep{zhang2024mavismathematicalvisualinstruction} and Math-PUMA \citep{zhuang2024mathpumaprogressiveupwardmultimodal}, utilized multi-stage (3-4) training pipelines with large-scale synthetic data, but faced challenges in computational efficiency and scalability to broader mathematical domains. These methods often rely heavily on synthetic data generation, limiting practical applicability. Thus, prior efforts remain computationally expensive or exhibit inconsistent performance in dense visual mathematical reasoning, particularly due to tackling these tasks in isolation. \proposedmodel\ addresses these challenges by advancing visual comprehension-infused symbolic reasoning in a computationally efficient framework.

\section{Conclusion}
In this paper, we introduced a novel framework, \proposedmodel, supplemented by a new dataset, \newdataset. This approach aims to enhance visual spatial comprehension and infuse it into the sound symbolic reasoning capabilities of MSLMs across a diverse range of visual and textual integrations, specifically addressing geometric mathematics problems. Empirical results reinforce that \proposedmodel\ surpasses contemporary baselines, thereby representing a significant advance in developing medium-sized visual mathematics solvers, which can proficiently interpret complex visual mathematical problems and make sound logical deductions.

\section{Limitations}
The proposed dataset, \newdataset\, is derived from an existing dataset, which necessitates reliance on the authors of the source dataset to establish the quality standards applied. Further, the efficacy of the reasoning-enhanced solver, \mrm, is contingent upon the outcomes generated by the \mcm. Therefore, any errors originating from the \mcmfull\ may propagate to the \mrm, potentially making an adverse impact, despite the mitigation strategies presented in this work.

Additionally, it is important to note that the dataset is exclusively based on the English language, thereby limiting our ability to assess the applicability and effectiveness of the proposed methodology in non-English contexts, particularly for low-resource languages.

\section{Ethical Considerations}
The proposed dataset, referred to as \newdataset\, represents an expansion of an existing publicly available dataset and will be released publicly in accordance with the source dataset's guidelines. All data annotators involved in this study were compensated fairly for their contributions. We recognize the lack of women's representation in \newdataset\ annotations as a reflection of the broader global gender disparity in research and innovation; however, the authors do not endorse this perspective. We are committed to promoting gender and racial equality within research fields. 

Furthermore, we clarify that no generative AI tools were utilized in the creation of this content, aside from those used for spell-checking and grammar correction. We emphasize that the scope of \newdataset\ is strictly dedicated to scientific research purposes. The selection of GPT-4o for data augmentation was a decision made based on its popularity, and the authors do not express bias toward any specific commercial large language model.

\section*{Acknowledgments}
Tanmoy Chakraborty acknowledges the support of Microsoft Research Grant, Azure AI Credits Grant from Microsoft's Accelerating Foundation Models Research (AFMR) initiative, Google GCP Grant, and Rajiv Khemani Young Faculty
Chair Professorship in Artificial Intelligence.

\bibliography{paper}

\appendix

\section{Data Statistics and Quality}
\label{App:data}

\subsection{MathVerse.} MathVerse provides a diverse range of visual infusion in geometric mathematical problems comprising multi-choice and free-form style Q/A pairs. It serves as a contemporary benchmark for evaluating MSLMs with respect to their visual mathematical reasoning capabilities. The dataset was derived from various publicly accessible datasets and subsequently transformed to incorporate varied degrees of visual infusion in mathematical inquiries. 

The construction of the questions within MathVerse posits that a multiple-choice geometry problem typically comprises three components in addition to the answer choices -- descriptive information, an implicit property, and an essential condition. Descriptive information pertains to the directly observable and distinctly represented elements within the diagram, such as the presence of geometric shapes or the intersection points of functions. The implicit property necessitates a higher level of visual perception to infer from the diagram, signifying critical visual conditions for problem-solving. This includes aspects such as the parallelism and perpendicularity of lines, the similarity and congruence of triangles, as well as the classification and periodicity of functions. The essential condition represents specific numerical or algebraic measurements. By leveraging these three properties along with the accompanying diagram, MathVerse presents six versions of each problem -- Text-Dominant, Text-Only, Text-Lite, Vision-Intensive, Vision-Dominant, and Vision-Only. Notably, the Vision-Only configuration presents the mathematical problem solely in visual modality, devoid of any textual information.

\subsection{\newdataset\ Statistics.}
In the construction of the \newdataset, this study focuses exclusively on 2D and 3D geometric problems, deliberately excluding issues related to functions to ensure the dataset is homogenized for geometry-related purposes. Additionally, we consider only five visual modalities derived from the source dataset: Text-Dominant, Text-Lite, Vision-Intensive, Vision-Dominant, and Vision-Only. The Text-Only modality has been excluded, as it does not align with the multimodal focus of this research. During the fine-tuning phases, a total of 2,260 instances were considered across all settings, while 500 instances were allocated for testing purposes in the results presented herein.

To ensure the quality of the augmented data, we exclusively consider the \rcfull\ (r) in which the model's final answer aligns with the ground truth during the generation process. This approach resulted in the selection of 1,690 high-quality training instances for the \mrm\ and the (SFT + data augmentation) baseline fine-tuning. In order to facilitate the model's learning of a consistent \rcfull, we utilize the \rcfull\ (r) derived from the Text-Dominant setting across all ranges of visual infusion. This strategy provides the model with the maximum input necessary for the generation of the \rcfull.

\subsection{Commentaries on Commercial LLM's Usage.}The recent advancements in open-source LLMs, exemplified by DeepSeek-R1 with publicly available code and weights, demonstrate performance comparable to that of commercial LLMs. This development significantly alleviates the accessibility challenges posed by proprietary systems such as GPT-4. The data augmentation capabilities of GPT-4 can be effectively substituted with the open-source DeepSeek-V3 \citep{deepseekai2024deepseekv3technicalreport} or LLaMA-4 \footnote{\url{https://ai.meta.com/blog/llama-4-multimodal-intelligence}}-Maverick.

\begin{figure*}[!t]
\noindent
\begin{tcolorbox}[title=, fonttitle=\small, boxrule=0.1mm, left=0mm, right=1mm]  
\small 
\textbf{Setting: Text Dominant}

\textbf{Prompt:}
    \begin{itemize}  
        \item Please generate the additional context for above question which may help in better comprehension of the problem. Context may include elaboration and simplification of the problem such that it becomes easier to solve it. You must not provide either the steps involve in the solution or the final solution to the problem
    \end{itemize}
\textbf{Question:}
    \begin{itemize}  
        \item As shown in the figure, in triangle ABC, it is known that angle A = 80.0, angle B = 60.0, point D is on AB and point E is on AC, DE parallel BC, then the size of angle CED is ()
        Choices: A: 40° B: 60° C: 120° D: 140° 
    \end{itemize}
\textbf{Image:}
\begin{center}    \includegraphics[width=0.15\textwidth]{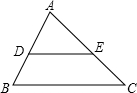} 
\end{center}
\textbf{Response - GPT4o:}
        \begin{itemize}  
        \item  
To better understand the problem, let's break down the given information and the relationships within the triangle:
1. **Triangle ABC**: This is the main triangle in the problem. We are given two angles:
   * Angle A = 80°
   * Angle B = 60°

2. **Point D on AB and Point E on AC**: These points divide the sides AB and AC respectively.

3. **DE is parallel to BC**: This is a crucial piece of information. When a line segment (DE) is parallel to one side of a triangle (BC) and intersects the other two sides (AB and AC), it creates similar triangles. In this case, triangle ADE is similar to triangle ABC.

4. **Angle Relationships**:

   * Since DE is parallel to BC, angle ADE is equal to angle ABC, and angle AED is equal to angle ACB due to the properties of parallel lines and corresponding angles.

5. **Finding Angle CED**:

   * To find angle CED, we need to use the properties of similar triangles and the given angles. Since DE is parallel to BC, angle CED will have a specific relationship with the angles in triangle ABC.

By understanding these relationships and properties, you can determine the size of angle CED using the given angles and the fact that DE is parallel to BC. This context should help in comprehending the problem and setting up the necessary steps to find the solution.
 
        \end{itemize}
\textbf{Response - LLaMA-4-Maverick:}
        \begin{itemize}  
        \item **Problem Context and Elaboration**

The given problem involves a triangle ABC with specific angle measures and points D and E on sides AB and AC, respectively. The task is to determine the size of angle CED given that DE is parallel to BC.

**Given Information:**
*   Triangle ABC with known angles: angle A = 80.0 degrees and angle B = 60.0 degrees.
*   Point D lies on side AB.
*   Point E lies on side AC.
*   DE is parallel to BC.

**Objective:** To find the measure of angle CED.

**Relevant Geometric Principles:**
*   The sum of the interior angles of a triangle is always 180 degrees.
*   When two lines are parallel, corresponding angles are equal.

**Understanding the Figure:**
Since DE is parallel to BC, corresponding angles formed by the transversal through points A, E, and C will be equal. This implies that angle AED is equal to angle ACB due to the corresponding angles property.

**Calculating Angle ACB:**
Using the fact that the sum of the interior angles of a triangle is 180 degrees, we can calculate angle ACB as follows:

angle A + angle B + angle C = 180 degrees
80 degrees + 60 degrees + angle C = 180 degrees
angle C = 180 degrees - 140 degrees
angle C = 40 degrees

**Implication for Angle CED:**

Since DE is parallel to BC, angle AED = angle ACB = 40 degrees. Knowing this, we can use the property that the sum of the angles on a straight line is 180 degrees to find angle CED.
angle AED + angle CED = 180 degrees
Substituting the known value of angle AED:
40 degrees + angle CED = 180 degrees
This sets the stage for calculating angle CED.

The problem requires using the properties of parallel lines and the angles within a triangle to solve for angle CED.
        \end{itemize}
\end{tcolorbox}
\caption{A qualitative analysis of \spcfull\ generational capabilities of GPT4o and Gemini-2.5-Pro.}
\label{fig: QA_SC_llms}
\end{figure*}

In Figure \ref{fig: QA_SC_llms}, we present a qualitative analysis of GPT4o and LLaMA-4-Maverick in generating \spcfull. Analysis reveals that both models demonstrate an impressive ability to interpret the visual math problem and translate it into a structured, conceptual breakdown. They effectively identify the core geometric components, such as the triangle's given angles and the crucial relationship created by the parallel lines. The models showcase two distinct yet valuable descriptive styles. GPT4o excels at providing a comprehensive overview of the geometric properties, correctly identifying concepts like similar triangles and corresponding angles, which lays a robust theoretical foundation. On the other hand, LLaMA-4-Maverick adopts a more direct, pedagogical approach, outlining a clear, step-by-step path to the solution by highlighting the key principles needed.




\section{Extended Results}
\label{App:extres}

\subsection{Performance Across Benchmarks}

In the experiment, the selected categories of data for Geometry 3K, MathVista, and MathVision were carefully defined to ensure thorough Geometry-based visual math analysis. For Geometry 3K, the entire test set comprised 610 samples, allowing for a comprehensive exploration of geometry-related concepts. In the case of MathVista, the data focused specifically on metrics associated with angle measurement within the subject of metric geometry - angle. Meanwhile, MathVision was centered around the task of geometry problem-solving, emphasizing language proficiency in English (geometry problem solving and language - english). The detailed results are presented in Table \ref{tab:benchcom}.

\begin{table*}[t!]
\centering
\small
\begin{tabular}{lcccccc}
\toprule

\textbf{Benchmarks}& \textbf{Default} &\textbf{With \mcm} &\textbf{Improvements(in \%)}& \textbf{Samples Size}\\
\midrule
MathVision & 15.61&	19.08&	3.47 & 173\\
MathVista & 5.48&	6.16&  0.68  &146\\
Geometry3K & 8.82&	14.81&	5.99 & 601\\
\bottomrule
\end{tabular}
\caption{(Detail results) Performance of LLaVA-NeXT-34B-based \mcm\ alogn with default LLaVA-NeXT-34B-based solver across three other benchmarks- MathVision, MathVista, and Geometry3K.}
\label{tab:benchcom}
\end{table*}

\subsection{Why \mcmfull\ Is Important} In this experiment, we evaluate the effectiveness of incorporating \spcfull\ into structured reasoning chains using evolutionary analysis. First, we fine-tune an additional \mcm\ variant, referred to as \mcm$_{text\_only}$, which utilizes the text from 'Text Only' setting, the maximal textual description available for a given question as an alternative \spc. This formulation enables a systematic assessment of the impact of providing detailed available textual problem descriptions across diverse visual contexts. We compare performance across three settings: the default solver with \mcm$_{text\_only}$, the default solver with \mcm, and the \proposedmodel, applied to two base architectures, LLaVA-NeXT-34B and Phi-4. The remaining experimental configurations align with those outlined in the primary study within the paper.

The findings presented in Table \ref{tab:mainablation} demonstrate two key observations. First, the inclusion of a detailed \spcfull, which encapsulates geometric elements and relationships, significantly improves the performance of the default solver when compared to utilizing only the maximal textual description of the problem. Second, the integration of \spcfull\ into \rcfull\ yields substantial enhancements across models and diverse visual-textual settings. These results underscore the effectiveness of the novel proposed methodology, visual-comprehension-aware structured symbolic reasoning (\proposedmodel), in augmenting the problem-solving capabilities of MSLMs.
\begin{table*}[t!]
\centering
\small

\resizebox{.95\textwidth}{!}{
\begin{tabular}{lcccccccccccc}
\toprule

\textbf{Models}& \multicolumn{2}{c}{\textbf{All}} &\multicolumn{2}{c}{\textbf{Text Dominant}} &\multicolumn{2}{c}{\textbf{Text Lite}}& \multicolumn{2}{c}{\textbf{Vision Intensive}}& \multicolumn{2}{c}{\textbf{Vision Dominant}}&  \multicolumn{2}{c}{\textbf{Vision Only}}\\
\midrule
\multicolumn{13}{l}{\textbf{(a): LLaVA-NeXT-34B as Base Model for both} \mcm\ \textbf{and} \mrm} \\
\multicolumn{13}{l}{\textbf{(b): Phi-4 as Base Model for both} \mcm\ \textbf{and} \mrm}\\
\midrule
& \textbf{(a)}& \textbf{(b)} & \textbf{(a)}& \textbf{(b)} & \textbf{(a)}& \textbf{(b)} &\textbf{(a)}& \textbf{(b)} &\textbf{(a)}& \textbf{(b)} &\textbf{(a)}& \textbf{(b)} \\
\midrule
\mcm$_{text\_only}$&18.0&29.2&25.0&	35.0&18.0&	30.0&21.0&	30.0&13.0&	30.0&13.0&	21.0\\
\mcm&19.4&30.4&\textbf{28.0}&	36.0&18.0&	34.0&19.0&	31.0&15.0&	26.0&17.0&	25.0\\
\proposedmodel&\textbf{23.0}&\textbf{43.6}&27.0&	\textbf{53.0}&\textbf{23.0}&	\textbf{48.0}&\textbf{28.0}&	\textbf{45.0}&17.0&	\textbf{41.0}&\textbf{20.0}&\textbf{31.0}\\
\bottomrule
\end{tabular}
}
\caption{Evolutionary Analysis for assessing the impact of \mcm.}
\label{tab:mainablation}
\vspace{-1mm}
\end{table*}

\subsection{Computational Efficiency Analysis}
The detailed experimental results are presented in Table \ref{tab:timecom}.

\begin{table*}[t!]
\centering
\small

\resizebox{.85\textwidth}{!}{
\begin{tabular}{lcccccc}
\toprule

\textbf{Setting}&\multicolumn{4}{c}{\textbf{Inference Time (in Sec)}}&\textbf{\#Tokens Generated} &\textbf{Accuracy (All)}\\
\midrule
& \textbf{run1} &\textbf{run2} &\textbf{run3}& \textbf{Average}& &\\

\midrule
Zeroshot &	10.17&	9.94&	10.20&	10.10&	73.2&	14\\
SFT + Data Augmentation&	68.27&	69.45	&70.51	&69.41	&438.05&	19.2\\
\mcm&	49.41&	49.01	&48.81&	49.08	&376.95&	19.4\\
\proposedmodel&	67.02&	66.66&	66.10	&66.59&	484.73	&23\\
\bottomrule
\end{tabular}
}
\caption{(Detail Results) Analysis on computational overhead in comparison to a single model approach. Bubble Size: Average \#tokens generated.}
\label{tab:timecom}
\vspace{-1mm}
\end{table*}

\subsection{Robustness Across Temperature Parameter}

The section demonstrates the robustness of the proposed methodology through the manipulation of the model parameter known as Temperature, which governs the randomness of the output. The findings presented in Table \ref{tab:tempsignificance} elucidate that the proposed method exhibits a high degree of robustness across various probabilistic vocabulary spaces, as influenced by the parameter temperature across three runs.

\begin{table*}[t!]
\centering
\small

\begin{tabular}{lcccccc}
\toprule

\textbf{Solver Models}& \textbf{All} &\textbf{Text Dominant} &\textbf{Text Lite}& \textbf{Vision Intensive}& \textbf{Vision Dominant}&  \textbf{Vision Only}\\
\midrule
SFT +  Data Augmentation&19.2&21.0&	23.0&	18.0&20.0&14.0\\
\midrule
Temp = 0 &23.0&27.0&23.0&28.0&17.0&20.0\\
Temp = 0.2&23.0&28.0&23.0&28.0&16.0&20.0\\
Temp = 1&23.2&29.0&23.0&28.0&17.0&19.0\\
Average & 23.1 & 28.0 & 23.0 & 28.0& 16.7 &19.7\\
\bottomrule
\end{tabular}
\caption{Significance of model's parameter Temperature. Rest of the settings are same as Table 1.}
\label{tab:tempsignificance}
\vspace{-1mm}
\end{table*}

\subsection{Commentaries on Closed-Source Models}
Table \ref{tab:supp-t3} presents the evaluation results for the MATHVERSE benchmark for closed-source models. These results are sourced from the original publication by \cite{zhang2024mathversedoesmultimodalllm} and are provided here purely for reference purposes. We observe that the exponential increase in the number of parameters in closed-source models enables them to outperform their open-source counterparts. Despite achieving the highest overall average performance of 63.1\%, GPT-4V remains suboptimal, highlighting the need for further efforts to enhance the performance of closed-source models. Notably, advancements in parameter fine-tuning methodologies are typically reliant on access to the model’s architecture. However, closed-source models restrict direct access to their architectures, limiting the application of such methodologies to open-source models. Given access to the architectural parameters, there is intense anticipation that strategies akin to \proposedmodel\ could similarly augment the capabilities of closed-source models.. It is important to note that the results encompass all six settings, including samples from geometry-specific problems as well as function-related samples.

\begin{table*}[t!]
\centering
\small
\centering

\begin{adjustbox}{width=\linewidth}
    \begin{tabular}{l|p{0.9cm}p{0.9cm}|p{0.9cm}p{0.9cm}|p{0.9cm}p{0.9cm}|p{0.9cm}p{0.9cm}|p{0.9cm}p{0.9cm}|p{0.9cm}p{0.9cm}|p{0.9cm}p{0.9cm}}
    \toprule
    \multirow{3}*{{Model}}    &\multicolumn{2}{c|}{{All}}
    &\multicolumn{2}{c|}{{\shortstack{\vspace*{0.1pt}\\Text\\\vspace*{0.2pt}\\Dominant}}} 
    &\multicolumn{2}{c|}{{\shortstack{\vspace*{0.1pt}\\Text\\\vspace*{0.2pt}\\Lite}}}
    &\multicolumn{2}{c|}{{\shortstack{\vspace*{0.1pt}\\Text\\\vspace*{0.2pt}\\Only}}}
    &\multicolumn{2}{c|}{{\shortstack{\vspace*{0.1pt}\\\ Vision\ \ \\\vspace*{0.2pt}\\Intensive}}}
    &\multicolumn{2}{c|}{{\shortstack{\vspace*{0.1pt}\\\ Vision\ \ \\\vspace*{0.2pt}\\Dominant}}}
    &\multicolumn{2}{c}{{\shortstack{\vspace*{0.1pt}\\\ Vision\ \ \\\vspace*{0.2pt}\\Only}}}\\
    \cmidrule{2-15}
    & CoT-E & \ Acc\   & CoT-E & Acc& CoT-E & Acc& CoT-E & Acc& CoT-E & Acc& CoT-E & Acc& CoT-E & Acc  \\
    \midrule
    \multicolumn{15}{c}{\textit{LLMs}}\\
    \cmidrule{1-15}
    ChatGPT &- &-  &51.3 & 33.3  & 38.5 & 18.9 & 51.3 & 33.3&- &-  &-& - & - &-\\
    GPT-4 &- &-  & 63.4 & 46.5 & 40.7 & 20.7 & 63.4 & 46.5&- &-  &-& - & - &- \\
    \cmidrule{1-15}
    \multicolumn{15}{c}{\textit{Closed-source MLLMs}}\\
    \cmidrule{1-15}
    Qwen-VL-Plus& 21.3 & 11.8 &26.0&15.7&21.2&11.1&25.2&14.5&18.5&9.0& 19.1 & 13.0&21.8& 10.0\\
    Gemini-Pro &35.3 & 23.5 & 39.8 & 26.3  & 34.7 & 23.5 & 44.5 & 27.3 & 32.0 & 23.0 & 36.8 & 22.3 & 33.3 & 22.2 \\
    Qwen-VL-Max &37.2 & 25.3 & 42.8 & 30.7  & 37.7 & 26.1 & 47.9 & 28.9 & 33.6 & 24.1 & 35.9 & 24.1 & 35.9 & 21.4 \\
    GPT-4V&\colorbox{backred!50}{54.4} &\colorbox{backred!50}{39.4} &\colorbox{backred!50}{63.1} &\colorbox{backred!50}{54.7} &\colorbox{backred!50}{56.6} &\colorbox{backred!50}{41.4} &\colorbox{backred!50}{60.3} &\colorbox{backred!50}{48.7} &\colorbox{backred!50}{51.4} &\colorbox{backred!50}{34.9} &\colorbox{backred!50}{50.8} &\colorbox{backred!50}{34.4} &\colorbox{backred!50}{50.3} &\colorbox{backred!50}{31.6}\\
    \bottomrule
    \end{tabular}
\end{adjustbox}
\caption{The below table is sourced from the original work presented by \cite{zhang2024mathversedoesmultimodalllm} for the reference purpose only. Mathematical evaluation on six problem versions in MATHVERSE's \textit{testmini} set. The highest accuracy for \colorbox{backred!50}{closed-source} MLLMs is marked in red.}
\label{tab:supp-t3}
\end{table*}

\section{Evaluator Modeling on \mcm\ Cont.}
\label{app:evaluator}

\subsection{Objective} The output of the \mcm\ module is contingent upon the quality and geometrical accuracy established within the visual figure generated by the process. Inadequate quality of the \spcfull\ (\spc) may adversely affect the \mrm\ core's performance. Hence, the primary aim of the Evaluator module is to systematically assess the quality of the generated \spc\ and to filter out any instances deemed unsatisfactory, thereby allowing only those that meet quality standards to proceed to the \mrm\ core. This process is intended to mitigate the detrimental effects of the \mcm\ module on the \mrm\ module. The Evaluator is positioned as an intermediary between the \mcm\ and \mrm\ modules. Various criteria may be interpreted to distinguish a good \spc\ from a bad one. In this context, we define the efficacy of \spc\ based on its utility within the \mrm\ core. A satisfactory \spcfull\ is characterized by its lack of adverse effects on the outputs generated by the \mrm\ module.

\subsection{Training Data} To achieve this objective, we first develop a training dataset to construct a robust evaluator model. Utilizing this training set of \newdataset, we create instances where the input comprises a question $q_{(t,v)}$ alongside its corresponding \spc\ produced by the \mcm\ module, with an appended inquiry: "Is the provided context faithful to the given context?" The expected output is a binary label of "yes" or "no," indicating whether the \spc\ has a detrimental impact on the output of the \mrm\ core. A label of 'no' denotes an adverse impact, while 'yes' signifies the absence of such an impact.

\subsection{Fine-tuning and Metrics} Subsequently, using the aforementioned training data, we proceed to fine-tune the evaluator models associated with both the base models LLaVA-NeXT-34B and Phi-4, respectively. The configurations utilized in this fine-tuning process are consistent with those employed during the fine-tuning of the \mcm\ and \mrm\ modules.
To quantify the impact of the evaluative model, we introduce the concept of the improvement ratio, presented in Equation \ref{eq:improvement_ratio}, which is defined as the overall improvement divided by the overall degradation associated with the utilization of \spc\ in conjunction with \mrm\, for the purpose of assessing the effectiveness of quality control. In this context, an improvement refers to the positive changes observed in the outcomes of the \mrm\ module resulting from the application of \spc\, while degradation signifies the adverse effects on the outcomes of the \mrm\ module when employing \spc\ as compared to the default settings.

\begin{equation}
\text{Improvement Ratio} = \frac{\text{Overall Improvements}}{\text{Overall Degradations}}  
\label{eq:improvement_ratio}
\end{equation} 

\subsection{Results} The experimental results presented in Table \ref{tab:evaluator} highlight the comparative performance of \mcm\ and \mcm\ with Eval model across two base models, LLaVA-NeXT-34B and Phi-4. Both approaches demonstrate noticeable improvements in "Accuracy (All) Improvements" over their respective baseline models, with Phi-4 showing slightly higher enhancement (+2.2\% versus +1.4\%). When analyzing sample-level metrics, \mcm\ with Eval yields fewer "Absolute Degradations" compared to \mcm\ alone, reflecting the refinement introduced by the evaluator. Notably, the improvement ratio is more pronounced for \mcm\ with Eval, particularly for Phi-4 (3.0 vs. 1.12), suggesting that the evaluator effectively balances improvements and degradations at the sample level. The utility tag further supports this observation, indicating higher 'yes' predictions with \mcm\ with Eval, underscoring its practical efficacy.

Findings also reveal a slight reduction in "Accuracy (All) Improvements" when incorporating the evaluator model, with LLaVA-NeXT-34B and Phi-4 showing decreases of 0.6\% and 1.4\%, respectively. This suggests that while the evaluator successfully minimizes "Absolute Degradations" and improves the overall imrpovement ratio, it comes at the cost of a decreased top-level accuracy. These findings highlight the need to optimize between maximizing accuracy and minimizing adverse impacts, striking a balance to ensure both reliability and fairness in model predictions.

\begin{table*}[t!]
\centering
\small

\resizebox{.75\textwidth}{!}{
\begin{tabular}{llcc}
\toprule

\textbf{Metric}&\textbf{Method}&\textbf{LLaVA-NeXT-34B} &\textbf{Phi-4}\\
\midrule

\multirow{2}{*}{Accuracy (All) Improvements} &  \mcm  & 1.4 & 2.2\\
  & \mcm\ + with Eval &0.8 & 0.8 \\
  \midrule
\multirow{2}{*}{Absolute Improvements} &  \mcm &61&57\\
  &  \mcm\ + with Eval &8 &6 \\
  \midrule
\multirow{2}{*}{Absolute Degradations} &   \mcm &54&51 \\
  &   \mcm\ + with Eval &4 &2 \\
  \midrule
\multirow{2}{*}{Improvement Ratio} &  \mcm &1.13 &1.12\\
  & \mcm\ + with Eval & 2.0 & 3.0\\
  \midrule
Utility Tag & Yes/500 &49 & 21 \\

\bottomrule
\end{tabular}
}
\caption{The comparison of \mcm\ with \mcm\ using Evaluator across different base models is presented here. "Accuracy (All) improvements" measures the enhancement over the default model, while "Absolute improvements" and "Absolute degradations" are assessed at the sample level. The utility tag indicates the 'yes' predictions made by the evaluator for the test set.}
\label{tab:evaluator}
\vspace{-1mm}
\end{table*}

\section{Experimental Setup Cont.}
We conduct experiments using LLaVA- and Phi-based architectures to evaluate the proposed method. To mitigate catastrophic forgetting \cite{shi2024continuallearninglargelanguage} due to limited training data, we freeze the LLM and vision tower components, fine-tuning only the MLP projector for both \mcm\ and \mrm\ for LLaVA-based model. Similarly, for Phi-4, we fine-tune lora adapter and vision encoder. All experiments are conducted once on same default random seed settings.

\subsection{Hyperparamters}
\label{App:hyperparameter}
In this section, we details all the hyperparameters to ensure reproducibility of the results. LLaVa-based models consist of three primary components: a large language model (LLM) backbone and a vision tower followed by a multilayer perceptron (MLP) adapter coneecting the two. In our approach, we exclusively finetune the MLP adapter while keeping the other two components fixed. The fine-tuning process is conducted over three epochs across various methods. Our computational infrastructure is comprised of a dual NVIDIA A100\footnote{\url{https://www.nvidia.com/en-in/data-center/a100/}} GPU setup. The detailed hyperparameters for LLaVA-NeXT-34B and Phi-4 finetuning are presented in Table \ref{tab:llavahyper} and Table \ref{tab:phihyper}, respectively. The peak GPU usage for training either core of the \proposedmodel\ was approximately 96 GPU hours, utilizing a shared compute environment.

\subsection{Prompts}
\label{App:prompts}
\begin{figure*}[!t]
\centering
\includegraphics[width=0.95\textwidth]{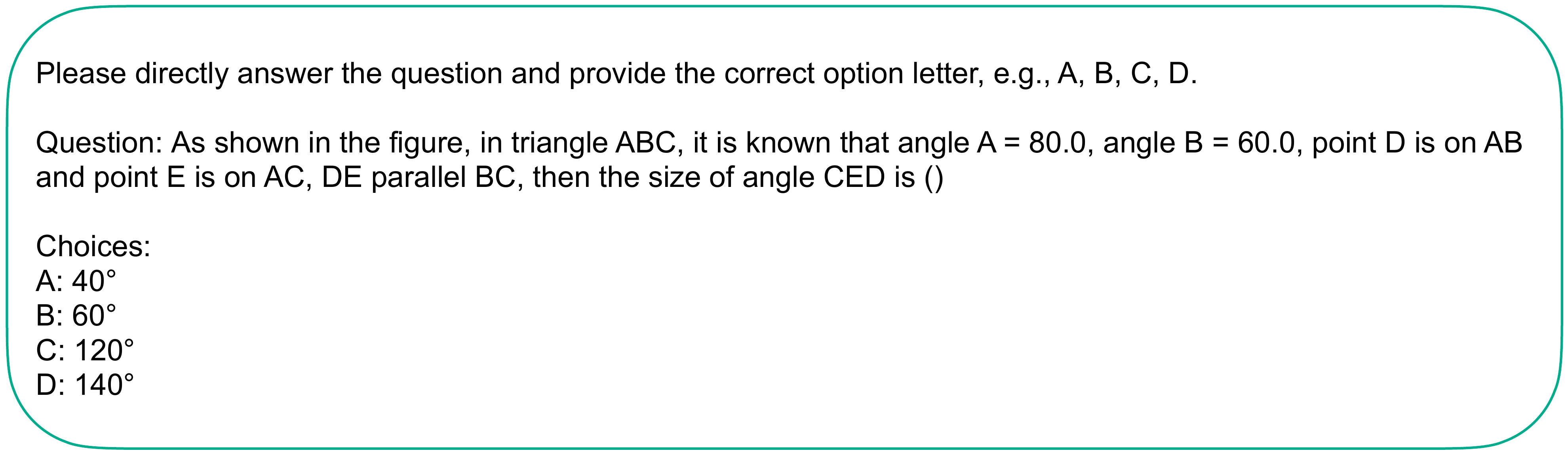}
  \caption{An example of prompt for Zero-shot inference across solver models for default performance benchmark along with an example problem.}
  \label{fig:inference_zeroshot}
\end{figure*}

\begin{figure*}[!t]
\centering
\includegraphics[width=0.95\textwidth]{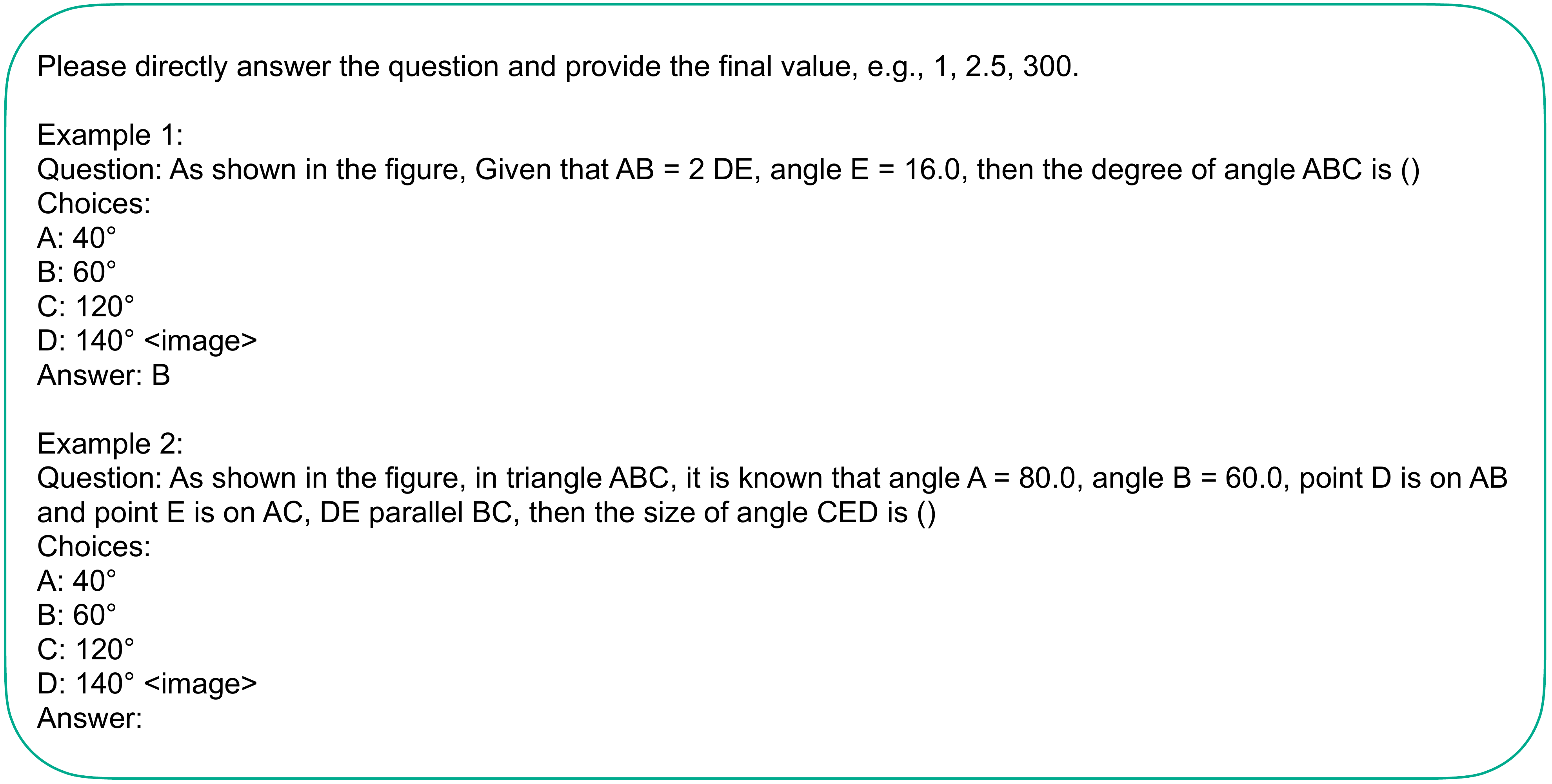}
  \caption{An example of prompt for one-shot ICL inference across solver models for default performance benchmark along with an example problem.}
  \label{fig:inference_icl}
\end{figure*}

\begin{figure*}[!t]
\centering
\includegraphics[width=0.95\textwidth]{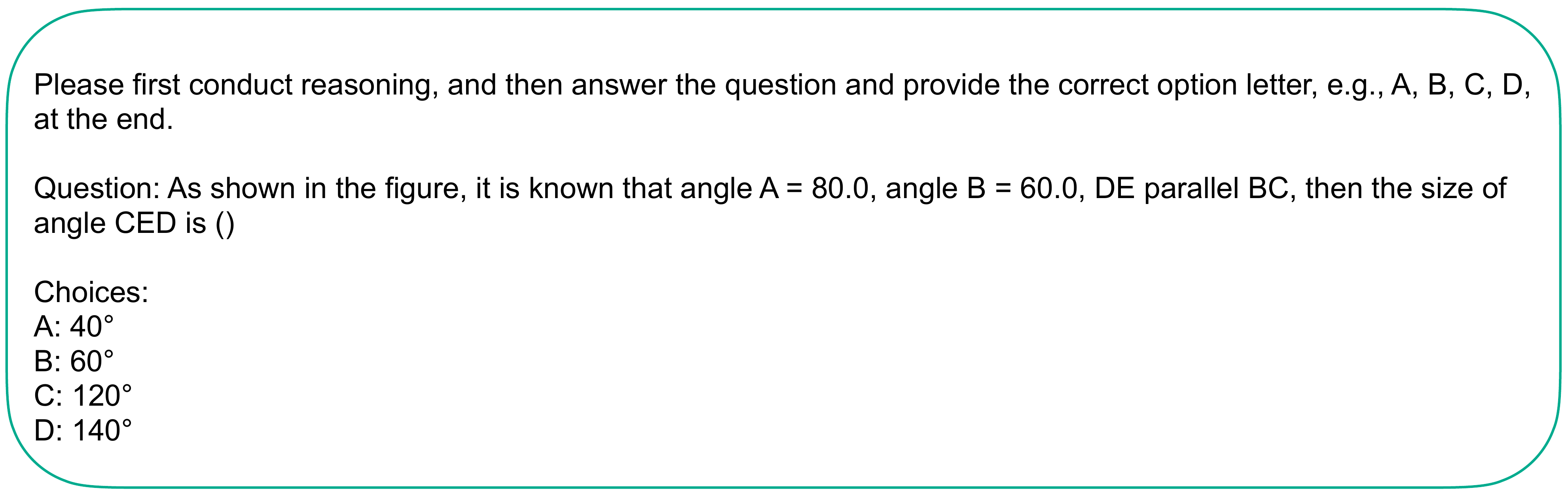}
  \caption{Prompt for one-shot CoT inference across solver models for default performance benchmark along with an example problem.}
  \label{fig:inference_cot}
\end{figure*}

\begin{figure*}[!t]
\centering
\includegraphics[width=0.95\textwidth]{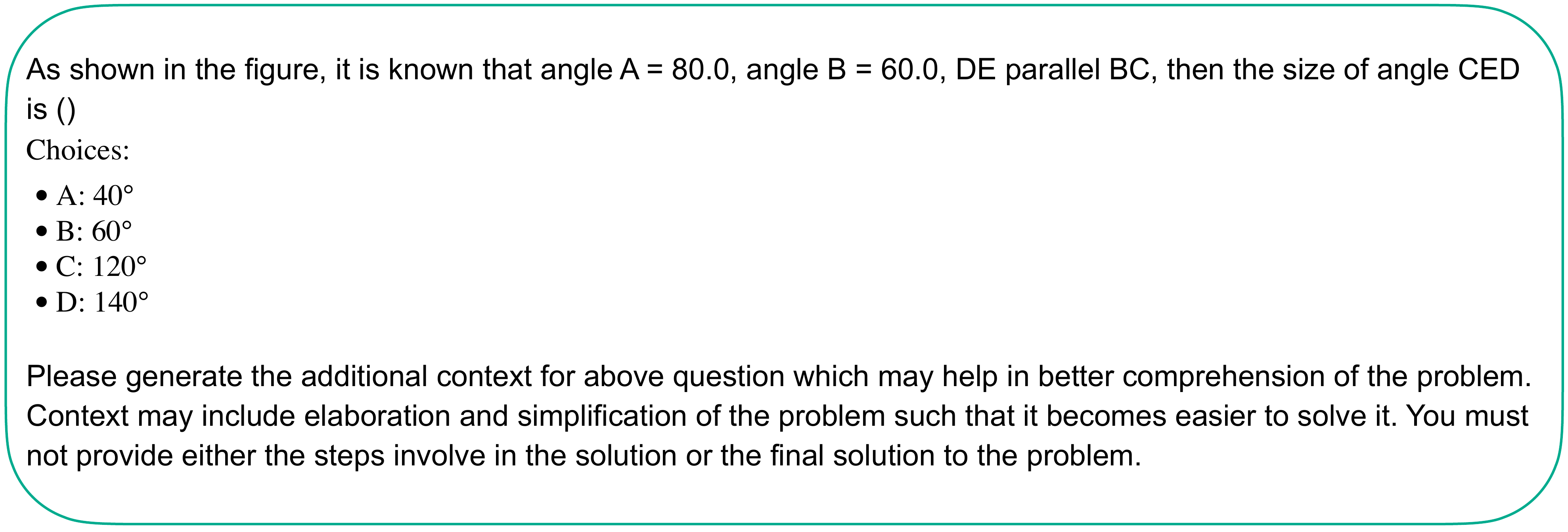}
  \caption{Prompt to generate perceptual interpretation from relatively larger model (GPT-4o): Data augmentation. The problem image was provided along with this prompt as inputs to the model.}
  \label{fig:PI_gen}
\end{figure*}

\begin{figure*}[!t]
\centering
\includegraphics[width=0.95\textwidth]{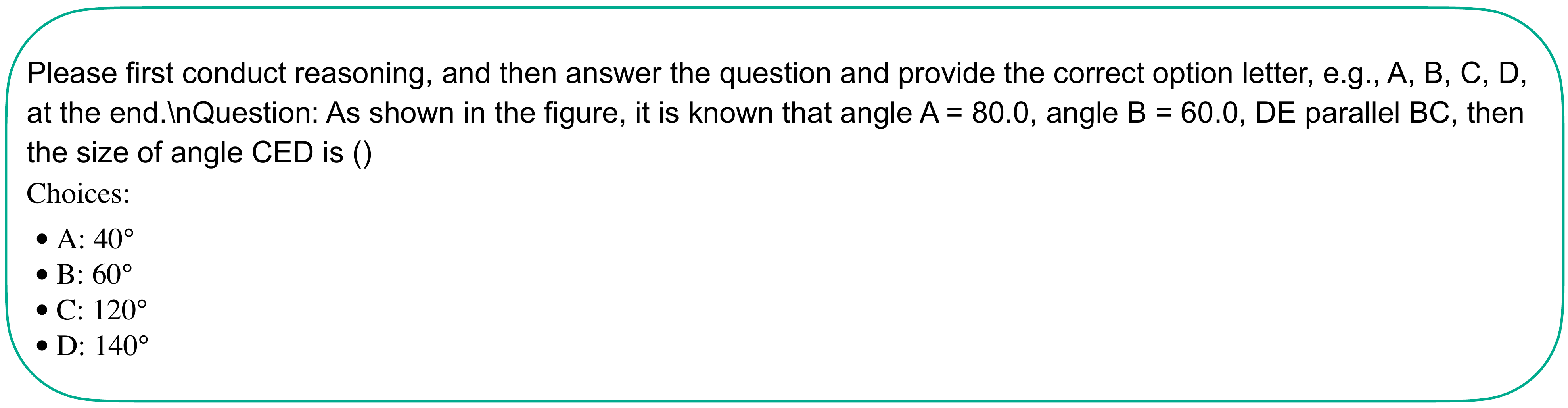}
  \caption{Prompt to generate reasoning chains from relatively larger model (GPT-4): Data augmentation. The problem image was provided along with this prompt as inputs to the model.}
  \label{fig:cot_gen}
\end{figure*}

\begin{figure*}[!t]
\centering
\includegraphics[width=0.95\textwidth]{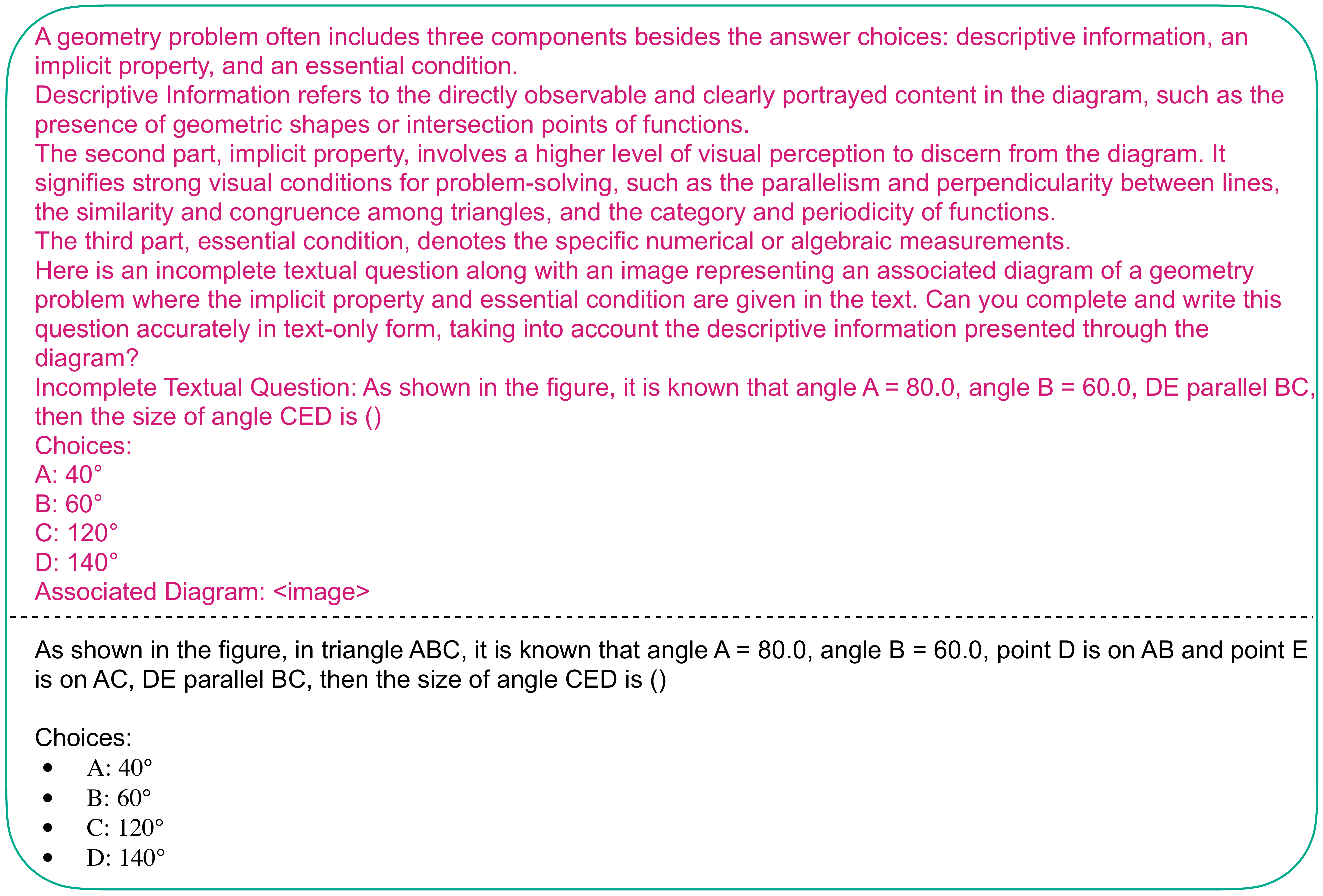}
  \caption{An instance of input (highlighted in magenta) and Output for fine-tuning objective of \mrm.}
  \label{fig:tr_ap1}
\end{figure*}

\begin{figure*}[!t]
\centering
\includegraphics[width=0.95\textwidth]{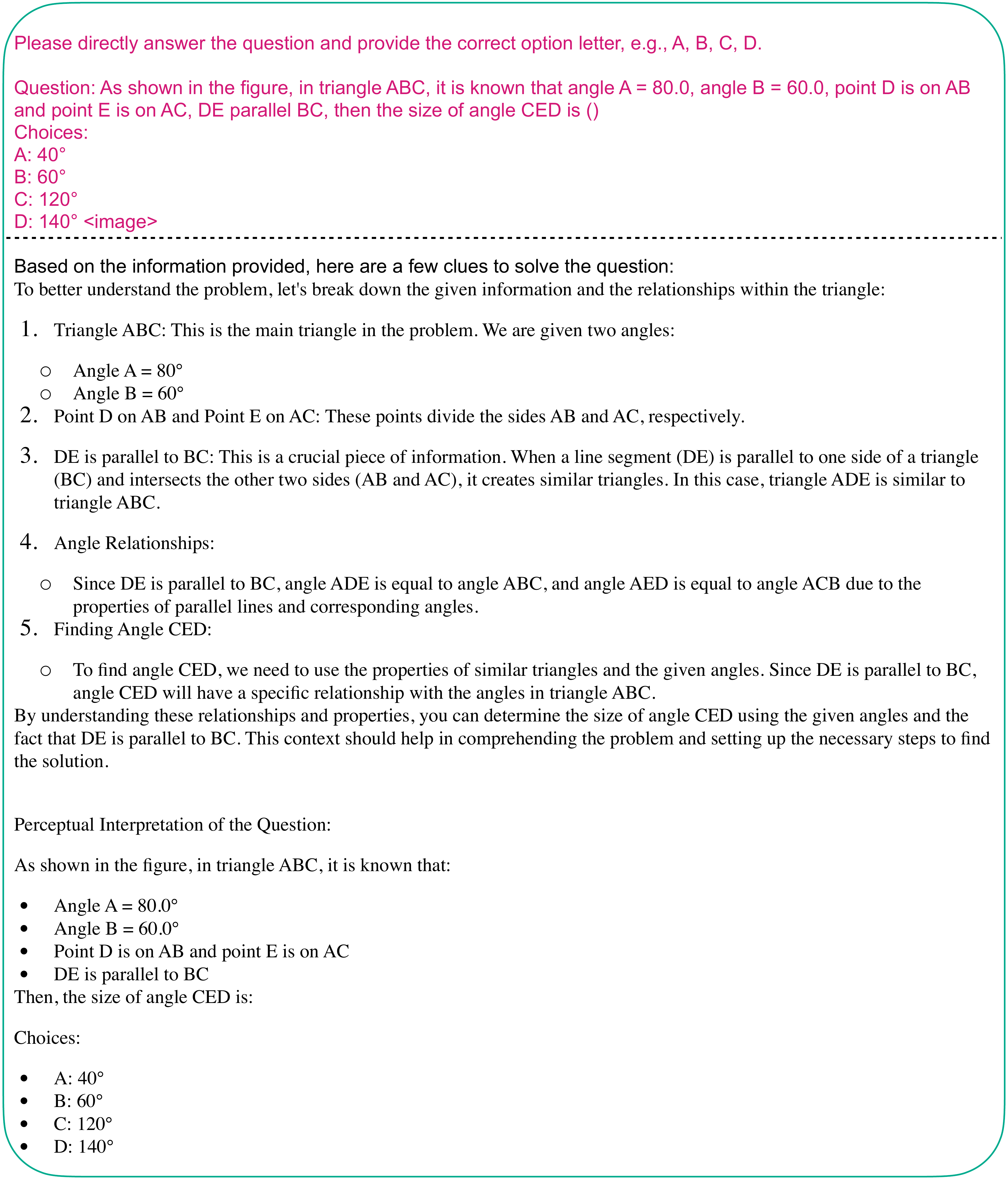}
  \caption{An instance of input (highlighted in magenta) and Output for fine-tuning objective of \mcm.}
  \label{fig:tr_ap2}
\end{figure*}

\begin{figure*}[!t]
\centering
\includegraphics[width=0.95\textwidth]{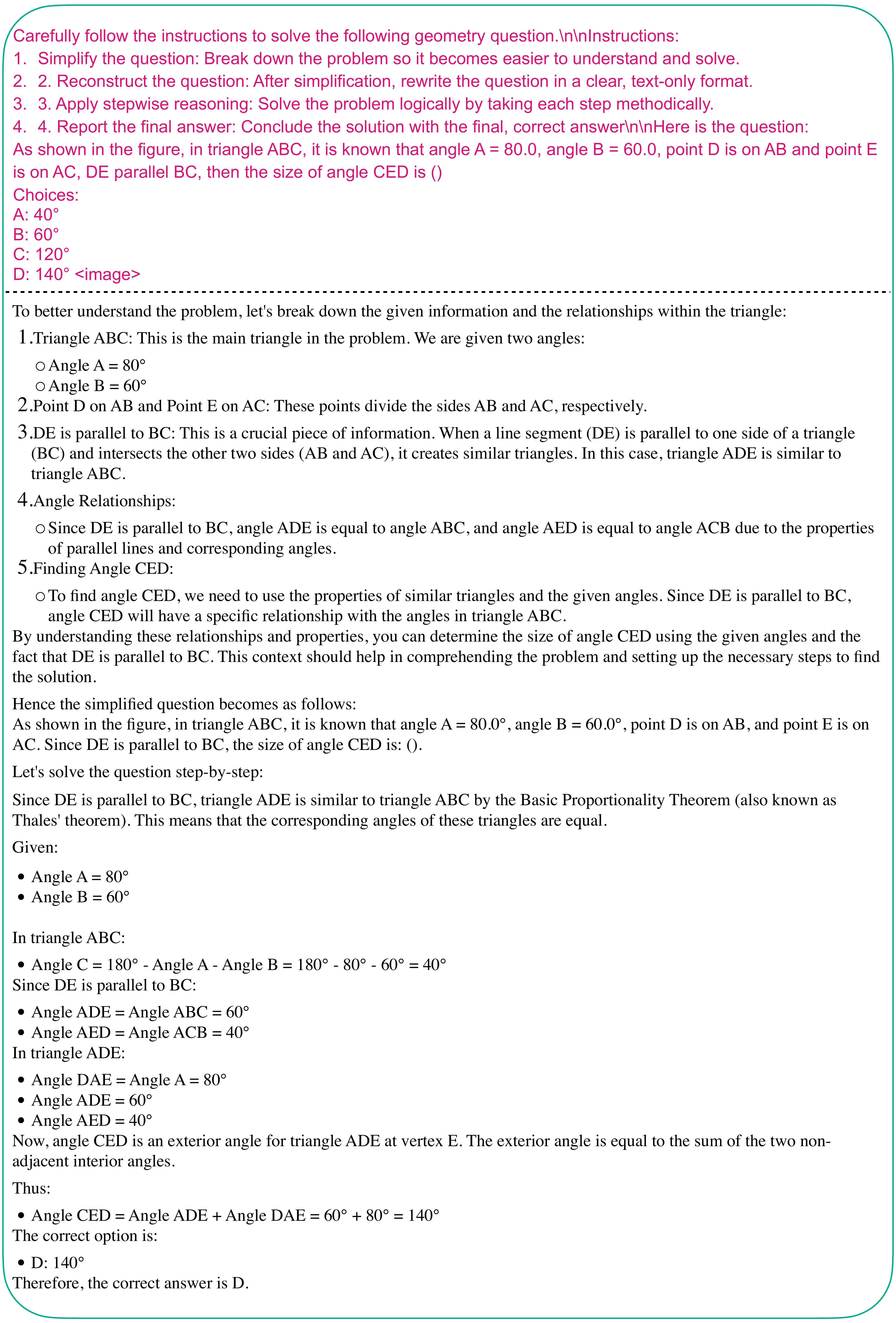}
  \caption{An instance of input (highlighted in magenta) and Output for fine-tuning objective of (SFT+Data Augmentation) baseline.}
  \label{fig:tr_ap3}
\end{figure*}

\begin{figure*}[!t]
\centering
\includegraphics[width=0.95\textwidth]{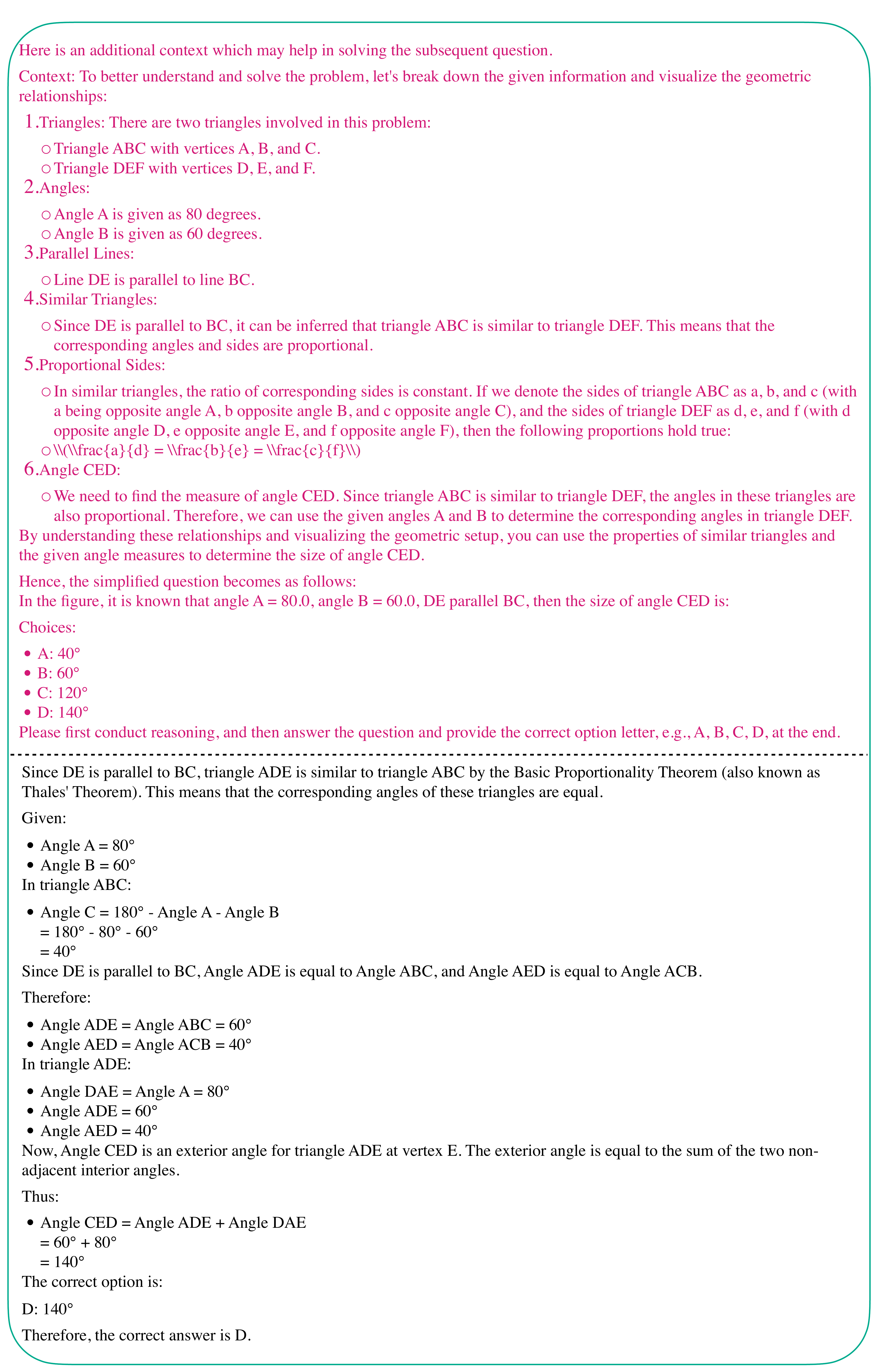}
  \caption{An instance of input (highlighted in magenta) and Output for fine-tuning objective of \proposedmodel.}
  \label{fig:tr_ap4}
\end{figure*}

\begin{figure*}[!t]
\centering
\includegraphics[width=0.95\textwidth]{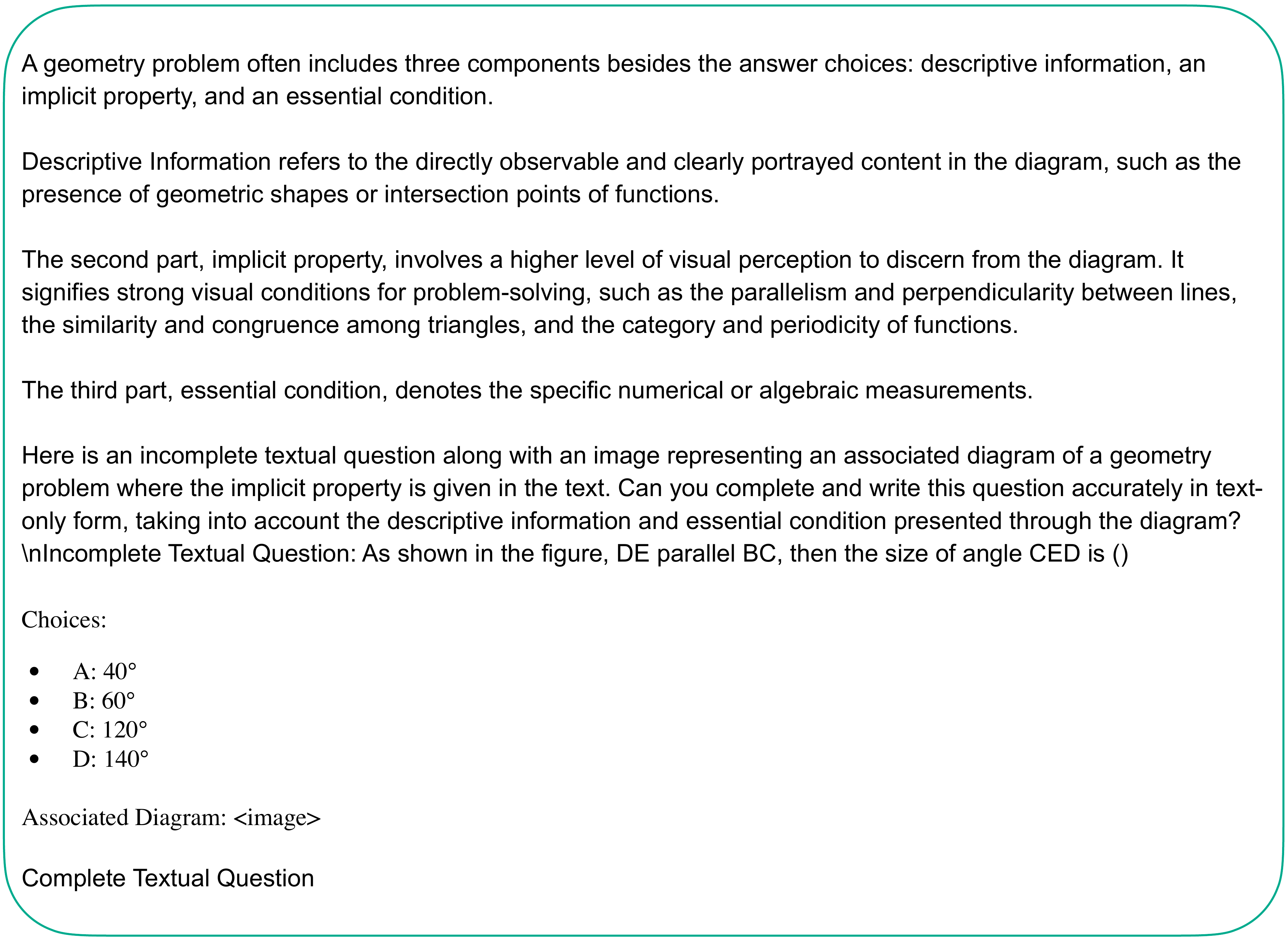}
  \caption{Inference prompt for $\mcm_{text\_only}$ based fine-tuned SC model along with an example problem.}
  \label{fig:infer_ap1}
\end{figure*}

\begin{figure*}[!t]
\centering
\includegraphics[width=0.95\textwidth]{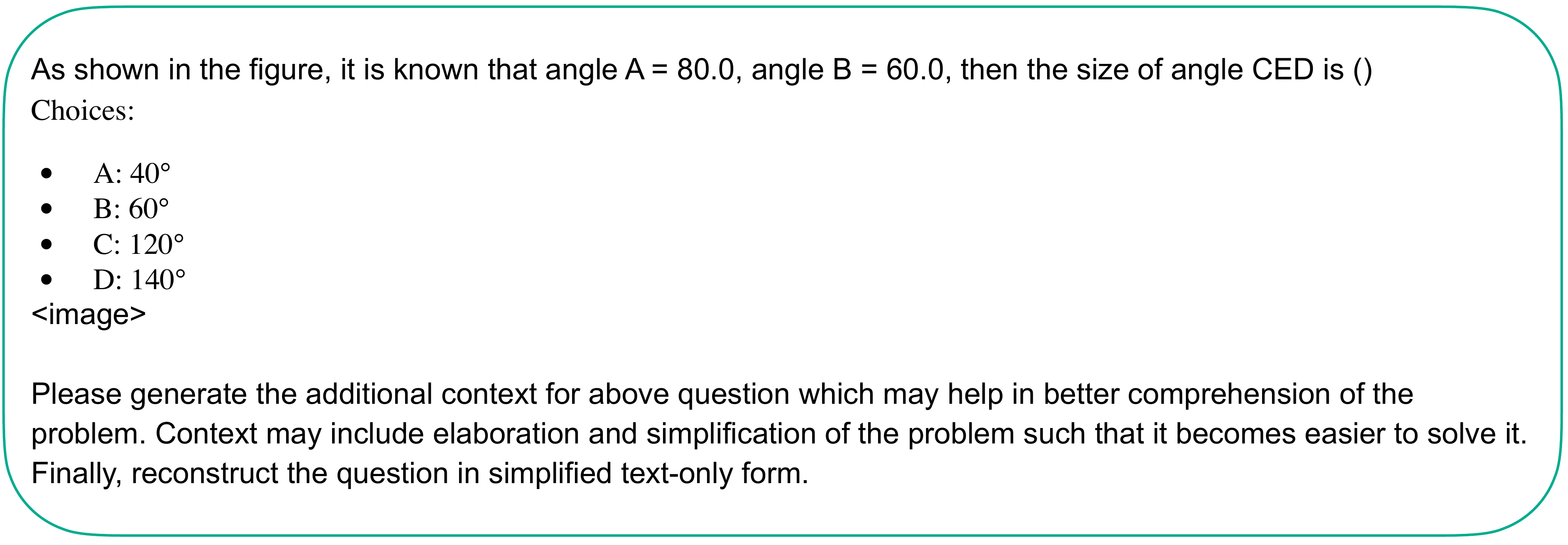}
  \caption{Inference prompt for \mcm\ based fine-tuned SC model along with an example problem.}
  \label{fig:infer_ap2}
\end{figure*}

\begin{figure*}[!t]
\centering
\includegraphics[width=0.95\textwidth]{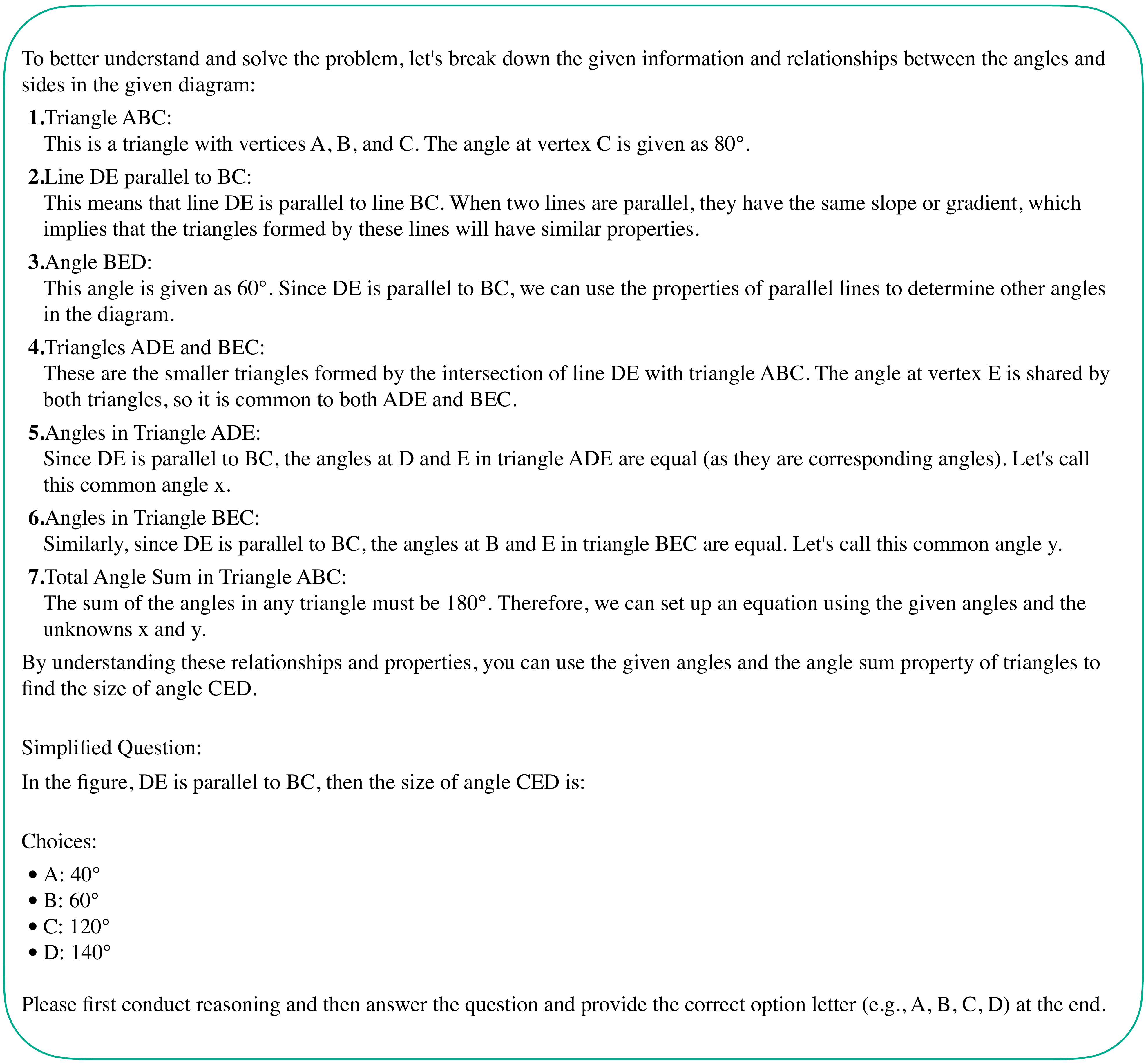}
  \caption{Inference prompt for default and \mrm\ based fine-tuned reasoning solver model considering PI from downstream SC model along with an example problem.}
  \label{fig:infer_solver_2_4}
\end{figure*}

In this study, we delineate the diverse prompts employed across various settings and fine-tuning procedures. The prompts utilized to evaluate zero-shot, in-context (one-shot), and chain-of-thought (one-shot) methodologies are depicted in Figures \ref{fig:inference_zeroshot}, \ref{fig:inference_icl}, and \ref{fig:inference_cot}, respectively. Additionally, a training data instance that encompasses an input and output for the objectives $\mcm_{text\_only}$, \mcm\, \proposedmodel, and the SFT+ data augmentation baseline is illustrated in Figures \ref{fig:tr_ap1}, \ref{fig:tr_ap2}, \ref{fig:tr_ap3}, and \ref{fig:tr_ap4}, respectively. Furthermore, the prompts deployed during the inference stages for the $\mcm_{text\_only}$ and \mcm\ fine-tuned study companion models are presented in Figures \ref{fig:infer_ap1} and \ref{fig:infer_ap2}, respectively. Concurrently, the prompts applied during the inference for the \mrm\ are shown in Figure \ref{fig:infer_solver_2_4}. The prompts utilized during the data augmentation phase to generate \spc\ and reasoning chain $r$ are illustrated in Figure \ref{fig:PI_gen} and Figure \ref{fig:cot_gen}, respectively.

Furthermore, we clarify that no generative AI tools were utilized in the creation of this content, aside from those used for spell-checking and grammar correction. We emphasize that the scope of \newdataset\ is strictly dedicated to scientific research purposes. The selection of GPT-4o for data augmentation was a decision made based on its popularity, and the authors do not express bias toward any specific commercial large language model.
\begin{table}[!htb]
    \centering
    
    \begin{tabular}{ll}
        \toprule
        \textbf{Hyperparameter} & \textbf{Value} \\
        \midrule
        Batch Size              & 16 \\
        Gradient Accumulation   & 16 \\
        Learning Rate           & 4.0e-5\\
        Weight Decay            & 0.01 \\
        Optimizer               & AdamW \\
        Scheduler               & Linear\\
        Warmup Steps / Ratio    & 50 steps\\
        Max Sequence Length     & 32768 \\
        Mixed Precision         & bf16 \\
        \midrule
        Inference: Model Parameters\\
        \midrule
        temperature & 1.0\\
        top\_p & 1.0 \\
        top\_k & 50 \\

        \bottomrule
    \end{tabular}
 \caption{Phi4 Hyperparameters.}
     \label{tab:phihyper}
\end{table}

\begin{table}[!htb]
    \centering
     
    \begin{tabular}{ll}
        \toprule
       \textbf{Hyperparameter} & \textbf{Value} \\
        \midrule

        Batch Size              & 1 \\
        Gradient Accumulation   & 1 \\
        Learning Rate           & 1e-5 \\
        Weight Decay            & 0.0 \\
        Optimizer               & AdamW \\
        Scheduler               & Cosine \\
        Warmup Ratio            & 0.03 \\
        Max Sequence Length     & 32768 \\
        Mixed Precision         & bf16 \\
        \midrule
        Inference:Model Parameters\\
        \midrule
        temperature & 0.7\\
        top\_p & 0.8 \\
        top\_k & 20 \\
        repetition\_penalty & 1.05 \\
        \bottomrule
    \end{tabular}
    \caption{LlaVa Next 34B Hyperparameters.}
     \label{tab:llavahyper}
\end{table}

\end{document}